# Reasoning about the Transfer of Control

**Wiebe van der Hoek** WIEBE.VAN-DER-HOEK@LIV.AC.UK
**Dirk Walther** DWALTHER@LIV.AC.UK
**Michael Wooldridge** MJW@LIV.AC.UK
*Department of Computer Science*
*University of Liverpool, UK*

## Abstract

We present DCL-PC: a logic for reasoning about how the abilities of agents and coalitions of agents are altered by transferring control from one agent to another. The logical foundation of DCL-PC is CL-PC, a logic for reasoning about cooperation in which the abilities of agents and coalitions of agents stem from a distribution of atomic Boolean variables to individual agents – the choices available to a coalition correspond to assignments to the variables the coalition controls. The basic modal constructs of CL-PC are of the form 'coalition $C$ can cooperate to bring about $\varphi$'. DCL-PC extends CL-PC with dynamic logic modalities in which atomic programs are of the form 'agent $i$ gives control of variable $p$ to agent $j$'; as usual in dynamic logic, these atomic programs may be combined using sequence, iteration, choice, and test operators to form complex programs. By combining such dynamic transfer programs with cooperation modalities, it becomes possible to reason about how the power of agents and coalitions is affected by the transfer of control. We give two alternative semantics for the logic: a 'direct' semantics, in which we capture the distributions of Boolean variables to agents; and a more conventional Kripke semantics. We prove that these semantics are equivalent, and then present an axiomatization for the logic. We investigate the computational complexity of model checking and satisfiability for DCL-PC, and show that both problems are PSPACE-complete (and hence no worse than the underlying logic CL-PC). Finally, we investigate the characterisation of *control* in DCL-PC. We distinguish between *first-order control* – the ability of an agent or coalition to control some state of affairs through the assignment of values to the variables under the control of the agent or coalition – and *second-order control* – the ability of an agent to exert control over the control that other agents have by transferring variables to other agents. We give a logical characterisation of second-order control.

## 1. Introduction

In recent years, there has been much activity in the development of logics for reasoning about the strategic and cooperative abilities of agents in game-like multi-agent systems. Coalition Logic (Pauly, 2001) and Alternating-time Temporal Logic (ATL) Alur, Henzinger, and Kupferman (2002) are perhaps the best-known examples of such work. These logics have been widely used as a base from which to investigate reasoning about cooperation in multi-agent systems (van der Hoek & Wooldridge, 2003; Jamroga & van der Hoek, 2004; Goranko & Jamroga, 2004).

Although they differ on details, the basic construct in both Coalition Logic and ATL is the *cooperation modality*, a construct that is written in ATL as $\langle\langle C\rangle\rangle\varphi$. The intended meaning of this expression is that the coalition $C$ can cooperate in such a way as to ensure that, no matter what the agents outside $C$ do, the property $\varphi$ becomes true. Another way to think about $\langle\langle C\rangle\rangle\varphi$ is as meaning that coalition $C$ has the *collective power* to ensure that $\varphi$. It is often assumed that powers are additive, in the sense that the powers of a coalition derive from the powers of coalition members, and that adding an agent to a coalition does not reduce the powers of that coalition. However, the *origin* of an individual agent's powers – that is, where these powers *derive from* – is rarely discussed in the cooperation logic literature.

One very natural interpretation for powers or abilities in computational systems arises from considering which system components have the ability to assign values to the variables making up the overall system state. Power, in this sense, equates to the ability to choose a value for a particular variable. Motivated by this observation, van der Hoek and Wooldridge developed CL-PC, a cooperation logic in which powers are specified by allocating to every agent a set of Boolean variables: the choices (and hence powers) available to a coalition then correspond to the possible assignments of truth or falsity that may be made to the variables under their control (van der Hoek & Wooldridge, 2005b). The CL-PC expression $\Diamond_C \varphi$ means that coalition $C$ can assign values to the variables under its control in such a way as to make $\varphi$ true. Van der Hoek and Wooldridge gave a complete axiomatization of CL-PC, and showed that the model checking and satisfiability problems for the logic are both PSPACE-complete; they also investigated how CL-PC could be used to characterise the closely related notion of control. However, one drawback of CL-PC is that the power structure underpinning the logic – the distribution of variables to agents – is assumed to be *fixed*, and hence coalitional powers are *static* in CL-PC.

Ultimately, of course, the assumption that powers are static is not realistic. For example, the explicit transfer of power and control is a fundamental component of most human organisations, enabling them to avoid bottlenecks with respect to centralised power and control. Moreover, in open environments, where agents join and leave a system at run-time, it may not be possible to know in advance which agents are to fulfill which roles, and so static power allocation schemes are simply not appropriate for such environments. If software agents are to be deployed in environments where power structures are dynamic, then it is important to consider the issues of representing and reasoning about them, and it is to this issue that we address ourselves in the present paper.

We study a variant of CL-PC that allows us to explicitly reason about dynamic power structures. The logic DCL-PC extends CL-PC with dynamic logic operators (Harel, Kozen, & Tiuryn, 2000), in which atomic programs are of the form $i \leadsto_p j$, which is read as 'agent $i$ gives control of variable $p$ to agent $j$'. The pre-condition of such a program is that variable $p$ is in agent $i$'s allocation of variables, and executing the program has the effect of *transferring variable p from agent i to agent j*. Thus the dynamic component of DCL-PC is concerned with *transferring power* in systems, and by using the logic, we can reason about how the abilities of agents and coalitions are affected by such transfers. Note that, as in conventional dynamic logic, atomic programs may be combined in DCL-PC with the usual sequential composition (';'), non-deterministic choice ('$\cup$'), test ('?'), and iteration ('$^*$') operations, to form complex programs. With these features, DCL-PC provides a rich framework through which to represent and reason about systems in which power/control is dynamically allocated.

In the remainder of the paper, following an introduction to the logic, we make four main contributions with respect to DCL-PC:

- First, in Section 2, we give two alternative semantics for the logic: a 'direct' semantics, in which models directly represent the allocation of propositional variables to the agents that control them, and a more conventional Kripke semantics. We prove that these two semantics are equivalent.

- Second, we give an axiomatization of DCL-PC in Section 3, and show that this axiomatization is sound and complete (with respect to both semantics).

- Third, we show in Section 4 that, despite the apparently additional expressive power provided by the dynamic component of DCL-PC, the satisfiability and model checking problems for DCL-PC are no more complex than the corresponding problems for CL-PC (van der Hoek & Wooldridge, 2005b): they are both PSPACE-complete.

- Fourth, we distinguish between *first-order control* and *second-order control* in Section 5. While first-order control, as introduced and studied by van der Hoek and Wooldridge (2005b), is the ability to control some state of affairs by assigning values to variables, second-order control is the ability of an agent to exert control over the ability of other agents to control states of affairs. Agents and coalitions can exercise second-order control by transferring variables under their control to other agents. After informally discussing and introducing second-order control, we develop a logical characterisation of it, in the sense that we characterise the formulae over which an agent has second-order control.

We conclude in brief with some comments on related work and conclusions. Note that we omit a detailed introduction to cooperation logics and in particular the motivation behind CL-PC, as this was done by van der Hoek and Wooldridge (2005b).

## 2. The Logic DCL-PC

In this section, we define the logic DCL-PC.

### 2.1 An Informal Introduction

We begin with an informal introduction; readers who are familiar with both CL-PC and dynamic logic may wish to skim or skip completely this introductory section.

As we noted earlier, DCL-PC extends the logic CL-PC, and we begin by briefly reviewing this logic. CL-PC is intended to allow us to reason about domains containing a collection of *agents*, and a collection of *propositional variables*; let $\mathbb{A} = \{1, \ldots, n\}$ denote the set of agents, and $\mathbb{P}$ denote the variables. It is assumed in CL-PC that each agent in the system *controls* some subset of the variables $\mathbb{P}$. To keep things simple, it is assumed that agents exercise unique control: every variable is controlled by exactly one agent, and so the variables $\mathbb{P}$ are partitioned among the agents $\mathbb{A}$. Where $i \in \mathbb{A}$ is an agent, we denote the variables under the control of $i$ by $\mathbb{P}_i$, so $\mathbb{P}_i \subseteq \mathbb{P}$. The *abilities* or *powers* of an agent in such a scenario correspond to the assignments of truth or falsity that it can make to the variables under its control. Figure 1 illustrates a typical scenario: we have four agents, $\mathbb{A} = \{1, 2, 3, 4\}$, and eight variables, $\mathbb{P} = \{p, q, r, s, t, u, v, w\}$. Agent 1 controls variables $p$ and $q$, ($\mathbb{P}_1 = \{p, q\}$), while agent 2 controls variable $r$, ($\mathbb{P}_2 = \{r\}$), and so on. In the scenario illustrated, variables $p, q, s$, and $w$ have the value

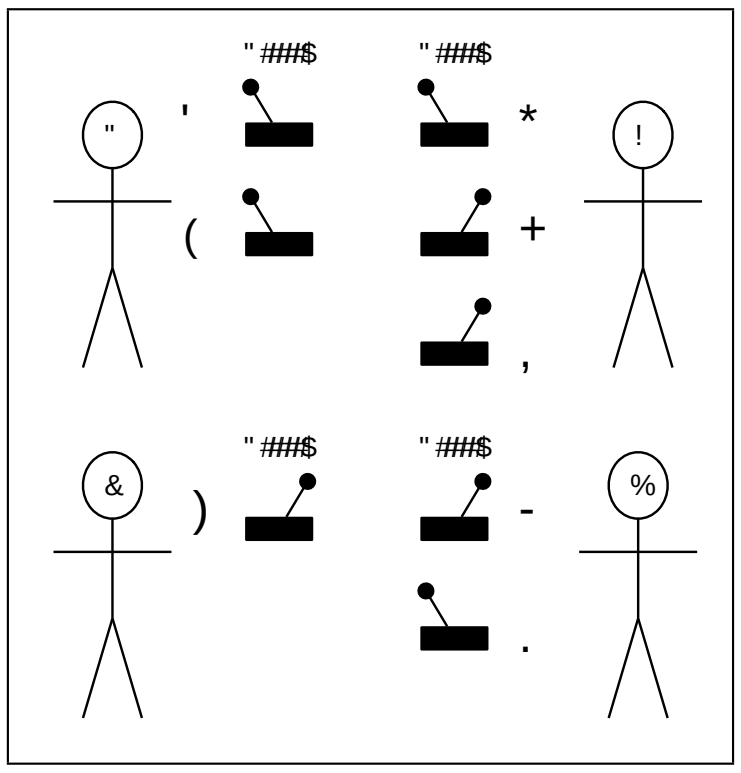

Figure 1: A typical scenario.

‘1’ (i.e., ‘true’), while all other variables have the value 0 (‘false’). The language of CL-PC is intended to allow us to represent and reason about such scenarios. To represent the values of variables, we use propositional logic, and so the following formula completely characterises the values of the variables in this scenario:

$$p \wedge q \wedge s \wedge w \wedge \neg r \wedge \neg t \wedge \neg u \wedge \neg v$$

Agents are able to change the value of the variables under their control, and to represent these abilities in CL-PC we use a *contingent ability operator* (van der Hoek & Wooldridge, 2005b): the expression $\Diamond_C \varphi$ means that, under the assumption that the world remains otherwise unchanged, the set of agents $C$ can modify the value of their variables so as to make $\varphi$ true. With respect to the scenario in Figure 1, for example, we have

$$\Diamond_{1,2}(p \wedge r \wedge \neg q).$$

This is because agent 1 can leave variable $p$ set at true while making variable $q$ false, while agent 2 makes variable $r$ true: this will result in the formula $p \wedge r \wedge \neg q$ being true.

The fact that no matter what coalition $C$ do, $\varphi$ will remain true is expressed by $\Box_C \varphi$. In the scenario in Figure 1, no matter what agent 1 does, $r$ will remain false (assuming again that no other agent acts). Thus we have:

$$\Box_1 \neg r.$$

As shown elsewhere (van der Hoek & Wooldridge, 2005b), and as defined below, other types of ability operators may also be defined.

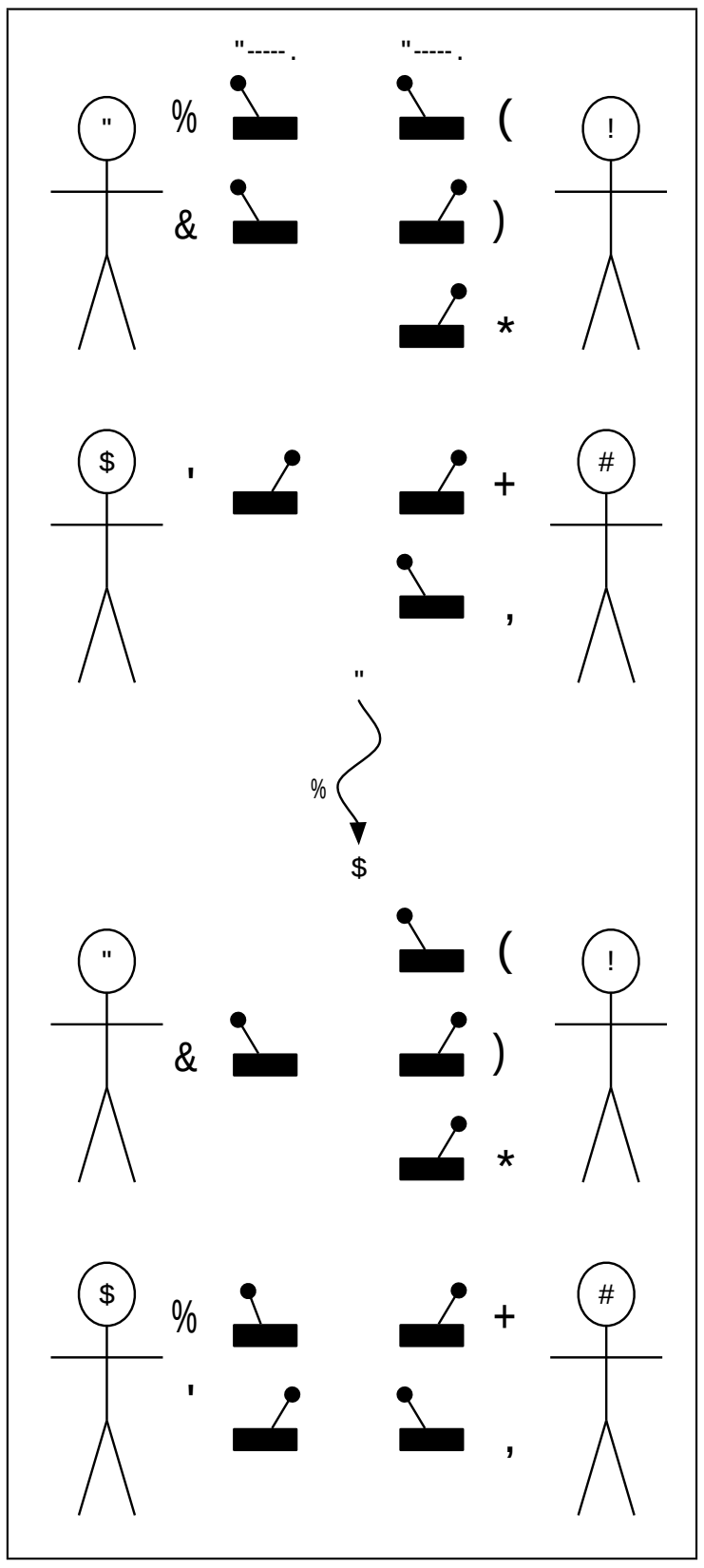

Figure 2: The effect of executing an atomic transfer program.

Thus far, the operators we have introduced have all been part of the CL-PC language. Let us now start to introduce the dynamic aspects of the language, specific to DCL-PC. First, we have the idea of an *atomic transfer program*, written $i \leadsto_p j$, meaning ‘agent $i$ transfers the power to choose a truth value for the variable $p$ to agent $j$’. Now, it will be possible to *execute* a program $i \leadsto_p j$ iff the variable $p$ is actually under the control of agent $i$. For example, with respect to Figure 1, the programs $1 \leadsto_p 2$ and $2 \leadsto_r 1$ are executable, while the program $1 \leadsto_r 2$ is not (since $r$ is not under the control of 1). The (fairly obvious) effect of executing the program $1 \leadsto_p 2$ is illustrated in Figure 2; note that *the actual* value *of the variable being transferred is unchanged by the transfer*.

In DCL-PC, we allow atomic programs to be combined together to make more complex programs using constructs from *dynamic logic*: ‘;’ (for sequential composition), ‘$*$’ (iteration), and ‘?’ (the ‘test’ operator). The simplest of these is sequential composition: for example, the program

$$1 \leadsto_p 2; 2 \leadsto_r 1$$

means that, first, agent 1 gives variable $p$ to 2, and then, agent 2 gives $r$ to 1. The operator $\cup$ is a *non-deterministic choice* operator. If $\tau_1$ and $\tau_2$ are transfer programs, then $\tau_1 \cup \tau_2$ means 'do *either* program $\tau_1$ or $\tau_2$'. The '$*$' operator is used to define iteration: the expression $\tau^*$ means 'execute the program $\tau$ zero or more times' (it is not defined exactly how many times $\tau$ will be executed). Finally, the '?' is used to perform *tests*. The program $\varphi?$ can be executed in a particular scenario only if the formula $\varphi$ is true of that scenario. To illustrate how these operator work, consider the following example programs.

$$p?; 1 \leadsto_p 2$$

This first program says 'if $p$ is true, then agent 1 gives $p$ to agent 2'. Now, since $p$ is true in the scenario in Figure 1, then this program can be executed from the scenario in Figure 1, and the net result is the same as the final scenario in Figure 2.

The following program uses non-deterministic choice, and essentially says 'agent 1 gives either $p$ or $q$ to 2'.

$$(1 \leadsto_p 2) \cup (1 \leadsto_q 2)$$

As is usual with dynamic logic, we can define the iteration and selection constructs used in conventional imperative programming languages from these basic program constructs. For example, the conventional programming construct

$$\texttt{while}\ \varphi\ \texttt{do}\ \tau$$

can be defined using the following transfer program construct (see, e.g., Harel et al., 2000):

$$(\varphi?; \tau)^*; \neg\varphi?$$

The next step is to see how transfer programs are incorporated with the ability constructs of CL-PC. To be able to refer to transfer programs and their properties from within the language of DCL-PC, we use the *dynamic operators* '$\langle\tau\rangle\varphi$' and '$[\tau]\varphi$'. These operators play the same role in DCL-PC that they play in conventional dynamic logic (Harel et al., 2000). Thus a formula $\langle\tau\rangle\varphi$ asserts that 'there exists a computation of program $\tau$, starting from the current situation, such that after $\tau$ has terminated $\varphi$ holds'. Note that $\langle\tau\rangle\varphi$ does not assert that $\tau$ is *guaranteed* to terminate, merely that it has *at least one* terminating computation. And moreover, it does not state that $\varphi$ is satisfied after *every* terminating computation of $\tau$; merely that there is *some* terminating computation that will end in a situation satisfying $\varphi$. Thus $\langle\tau\rangle$ acts as an existential quantifier over the computations of $\tau$. The operator $[\tau]\varphi$ is a universal quantifier over the computations of $\tau$. It asserts that after *every* terminating computation of $\tau$, the property $\varphi$ holds. Note that it does *not* assert that $\tau$ in fact has any terminating computations.

As an example of the use of these constructs, the following formula asserts that if agent 1 gives either $p$ or $q$ to 2, then 2 will be able to achieve $(p \vee q) \wedge r$.

$$[(1 \leadsto_p 2) \cup (1 \leadsto_q 2)]\Diamond_2(p \vee q) \wedge r$$

It is easy to see that this formula expresses a true property of the scenario in Figure 1: the program $(1 \leadsto_p 2) \cup (1 \leadsto_q 2)$ is executable in this scenario, and after it is executed, agent 2 will control variable $r$ and either variable $p$ or variable $q$. Agent 2 will thus be able to make $(p \vee q) \wedge r$ true.

To conclude this introductory section, consider the following more complex example. The following DCL-PC formula asserts that it is possible for agent $i$ to give away its variables to agent $j$, non-deterministically choosing one variable at a time, until agent *j* has the ability to achieve $\varphi$.

$$\langle \texttt{while} \neg\Diamond_j\varphi \texttt{ do } \bigcup_{p\in\mathbb{P}_i} i \rightsquigarrow_p j\rangle\top$$

### 2.2 Syntax

Formally, the language of DCL-PC is formed with respect to a (fixed, finite, non-empty) set $\mathbb{A}$ of agents, and a (fixed, finite, non-empty) set $\mathbb{P}$ of propositional variables. Figure 3 defines the syntax of DCL-PC. We use $\top$ as a logical constant for truth, '$\neg$' for negation, and '$\vee$' for disjunction. As usual, we define the remaining connectives of classical propositional logic as abbreviations:

$$\begin{aligned}
\bot &\hat{=} \neg\top \\
\varphi \wedge \psi &\hat{=} \neg(\neg\varphi \vee \neg\psi) \\
\varphi \rightarrow \psi &\hat{=} \neg\varphi \vee \psi \\
\varphi \leftrightarrow \psi &\hat{=} (\varphi \rightarrow \psi) \wedge (\psi \rightarrow \varphi).
\end{aligned}$$

Additionally, where $\Phi$ is a set of DCL-PC formulas, we write $\triangledown_{\varphi\in\Phi}\varphi$ to mean that exactly one member of $\Phi$ is true:

$$\triangledown_{\varphi\in\Phi}\varphi \;\hat{=}\; \bigvee_{\varphi\in\Phi}\varphi \wedge \bigwedge_{\varphi_1\neq\varphi_2\in\Phi} \neg(\varphi_1 \wedge \varphi_2).$$

Where $\Phi = \{\varphi_1, \varphi_2 \ldots \varphi_n\}$, we will also write $\varphi_1 \triangledown \varphi_2 \triangledown \cdots \triangledown \varphi_n$ for $\triangledown_{\varphi\in\Phi}\varphi$.

With respect to transfer programs, other constructs from conventional imperative programs may be defined as follows (Harel et al., 2000):

$$\begin{aligned}
\texttt{if } \varphi \texttt{ then } \tau_1 \texttt{ else } \tau_2 &\hat{=} ((\varphi?;\tau_1) \cup (\neg\varphi?;\tau_2)) \\
\texttt{while } \varphi \texttt{ do } \tau &\hat{=} ((\varphi?;\tau)^*;\neg\varphi?) \\
\texttt{repeat } \tau \texttt{ until } \varphi &\hat{=} \tau;(\neg\varphi?;\tau)^*;\varphi? \\
\texttt{skip} &\hat{=} \top? \\
\texttt{fail} &\hat{=} \bot?
\end{aligned}$$

Where there is no possibility of confusion, we will omit set brackets for cooperation modalities, for example writing $\Diamond_{1,2}$ rather than $\Diamond_{\{1,2\}}$. A DCL-PC formula containing no modalities is said to be an *objective* formula.

Let $\mathbb{P}(\varphi)$ denote the set of propositional variables occurring in DCL-PC formula $\varphi$, and let $\mathbb{A}(\varphi)$ denote the set of all agents named in $\varphi$ (i.e., $\mathbb{A}(\varphi)$ is the union of all the coalitions occurring in cooperation modalities in $\varphi$ and all the agents occurring in transfer programs in $\varphi$).

Although the $^*$ operator is useful to define programs succinctly, we will in fact see in Theorem 2 that it is superfluous, which essentially uses the fact that our set of atoms and agents are finite.

### 2.3 Direct Semantics

We now introduce the first of our two semantics for DCL-PC. We call this semantics a 'direct' semantics because it is directly based on the intuitive model that we introduced earlier: every agent has unique control of some set of propositional variables, with every variable being controlled by some agent.

Given a fixed, finite and non-empty set $\mathbb{A}$ of agents, and a fixed, finite and non-empty set $\mathbb{P}$ of propositional variables, we say an *allocation* of $\mathbb{P}$ to $\mathbb{A}$ is an indexed tuple $\xi = \langle\mathbb{P}_1, \ldots, \mathbb{P}_n\rangle$, where there is an indexed element $\mathbb{P}_i$ for each $i \in \mathbb{A}$, such that $\mathbb{P}_1, \ldots, \mathbb{P}_n$ forms a partition of $\mathbb{P}$ (i.e., $\mathbb{P} = \bigcup_{i \in \mathbb{A}} \mathbb{P}_i$ and $\mathbb{P}_i \cap \mathbb{P}_j = \emptyset$ for all $i \neq j \in \mathbb{A}$). The intended interpretation of an allocation $\xi = \langle \mathbb{P}_1, \ldots, \mathbb{P}_n \rangle$ is that $\mathbb{P}_i \subseteq \mathbb{P}$ is the set of propositional variables under agent $i$'s control. That is, agent $i$ has freedom to allocate whatever Boolean values it sees fit to the members of $\mathbb{P}_i$. Of course, we could have defined an allocation $\xi$ as a function $\xi : \mathbb{P} \to \mathbb{A}$, so that $\xi(p)$ denotes the agent controlling propositional variable $p$; there seems no particular reason for preferring one representation rather than the other, and so for consistency with the historical record, we will adopt the partition representation, as used by van der Hoek and Wooldridge (2005b).

| DCL-PC formulas: | | | |
|---|---|---|---|
| DCL | ::= | $\top$ | /* truth constant */ |
| | $\mid$ | $p$ | /* propositional variables */ |
| | $\mid$ | $\neg$DCL | /* negation */ |
| | $\mid$ | DCL $\vee$ DCL | /* disjunction */ |
| | $\mid$ | $\Diamond_C$DCL | /* contingent cooperative ability */ |
| | $\mid$ | $\langle\tau\rangle$DCL | /* existential dynamic operator */ |
| Transfer programs: | | | |
| $\tau$ | ::= | $i \leadsto_p j$ | /* $i$ gives $p$ to $j$ */ |
| | $\mid$ | $\tau;\tau$ | /* sequential composition */ |
| | $\mid$ | $\tau \cup \tau$ | /* non-deterministic choice */ |
| | $\mid$ | $\tau^*$ | /* iteration */ |
| | $\mid$ | DCL? | /* test */ |

Figure 3: Syntax of DCL-PC: $p \in \mathbb{P}$ is a propositional variable, $C \subseteq \mathbb{A}$ is a set of agents, and $i, j \in \mathbb{A}$ are agents.

Now, we say a *model* for DCL-PC is a structure:

$$\mathcal{M} = \langle \mathbb{A}, \mathbb{P}, \xi_0, \theta \rangle$$

where:

- $\mathbb{A} = \{1, \ldots, n\}$ is a finite, non-empty set of *agents*;
- $\mathbb{P} = \{p, q, \ldots\}$ is a finite, non-empty set of *propositional variables*;
- $\xi_0 = \langle \mathbb{P}_1, \ldots, \mathbb{P}_n \rangle$ is the *initial allocation* of $\mathbb{P}$ to $\mathbb{A}$, with the intended interpretation that $\mathbb{P}_i$ is the subset of $\mathbb{P}$ representing those variables under the control of agent $i \in \mathbb{A}$; and finally,
- $\theta : \mathbb{P} \to \{\mathrm{tt}, \mathrm{ff}\}$ is a *propositional valuation function*, which determines the initial truth value of every propositional variable.

Some additional notation is convenient in what follows. A *coalition* $C$ is a subset of $\mathbb{A}$, i.e., $C \subseteq \mathbb{A}$. For any such $C \subseteq \mathbb{A}$, we denote the *complement* of $C$, (i.e., $\mathbb{A} \setminus C$) by $\overline{C}$. We will write $\mathbb{P}_C$ for $\bigcup_{i \in C} \mathbb{P}_i$. For two valuations $\theta$ and $\theta'$, and a set of propositional variables $\Psi \subseteq \mathbb{P}$, we write $\theta = \theta' \pmod{\Psi}$ if $\theta$ and $\theta'$ differ at most in the propositional variables in $\Psi$, and we then say that $\theta$ and $\theta'$ are the *same modulo* $\Psi$. We will sometimes understand the model $\mathcal{M}$ to consist of a *frame* $\mathcal{F} = \langle \mathbb{A}, \mathbb{P}, \xi_0 \rangle$ together with a propositional valuation function $\theta$. Given a model $\mathcal{M} = \langle \mathbb{A}, \mathbb{P}, \xi_0, \theta \rangle$ and a coalition $C$ in $\mathcal{M}$, a *C-valuation* is a function:

$$\theta_C : \mathbb{P}_C \to \{\mathrm{tt}, \mathrm{ff}\}.$$

Thus a $C$-valuation is a propositional valuation function that assigns truth values to just the propositional variables controlled by the members of the coalition $C$. If $\mathcal{M} = \langle \mathbb{A}, \mathbb{P}, \xi_0, \theta \rangle$ with $\xi_0 = \langle \mathbb{P}_1 \ldots, \mathbb{P}_n \rangle$ is a model, $C$ a coalition in $\mathcal{M}$, and $\theta_C$ a $C$-valuation, then by $\mathcal{M} \oplus \theta_C$ we mean the model $\langle \mathbb{A}, \mathbb{P}, \xi_0, \theta' \rangle$, where $\theta'$ is the valuation function defined as follows

$$\theta'(p) \hat{=} \begin{cases} \theta_C(p) & \text{if } p \in \mathbb{P}_C \\ \theta(p) & \text{otherwise} \end{cases}$$

and all other elements of the model are as in $\mathcal{M}$. Thus $\mathcal{M} \oplus \theta_C$ denotes the model that is identical to $\mathcal{M}$ except that the values assigned by its valuation function to propositional variables controlled by members of $C$ are determined by $\theta_C$.

We define the *size* of a model $\mathcal{M} = \langle \mathbb{A}, \mathbb{P}, \xi_0, \theta \rangle$ to be $|\mathbb{A}| + |\mathbb{P}|$; we denote the size of $\mathcal{M}$ by *size*$(\mathcal{M})$.

### 2.4 Transfer Program Relations

To give a modal semantics to the dynamic logic constructs of DCL-PC, we must define, for every transfer program $\tau$ a binary relation $R_\tau$ over models such that $(\mathcal{M}_1, \mathcal{M}_2) \in R_\tau$ iff $\mathcal{M}_2$ is a model that may result from one possible execution of program $\tau$ from $\mathcal{M}_1$. We start by defining the relation $R_{i \leadsto_p j}$, for atomic transfer programs of the form $i \leadsto_p j$, i.e., agent $i$ gives control of propositional variable $p$ to agent $j$. Let $\mathcal{M} = \langle \mathbb{A}, \mathbb{P}, \xi_0, \theta \rangle$ and $\mathcal{M}' = \langle \mathbb{A}', \mathbb{P}', \xi_0', \theta' \rangle$ be two models with $\xi_0 = \langle \mathbb{P}_1, \ldots, \mathbb{P}_n \rangle$ and $\xi_0' = \langle \mathbb{P}_1', \ldots, \mathbb{P}_n' \rangle$. Then

$$(\mathcal{M}, \mathcal{M}') \in R_{i \leadsto_p j}$$

iff

1. $p \in \mathbb{P}_i$ (agent $i$ controls $p$ to begin with)
2. in case $i = j$:
   (a) $\mathcal{M} = \mathcal{M}'$ (agent $i$ gives $p$ to herself, with no change in the model)
3. in case $i \neq j$:
   (a) $\mathbb{P}_i' = \mathbb{P}_i \setminus \{p\}$ (agent $i$ no longer controls $p$ afterwards);
   (b) $\mathbb{P}_j' = \mathbb{P}_j \cup \{p\}$ (agent $j$ controls $p$ afterwards); and
   (c) all other components of $\mathcal{M}'$ are as in $\mathcal{M}$.

In order to define $\mathcal{M} \models^{\mathsf{d}} \varphi$, which means that $\varphi$ is true in $\mathcal{M}$ under the direct semantics, we need to be able to determine what the interpretation of an arbitrary program is, on $\mathcal{M}$; we define this below. Notice that executing an atomic transfer program has *no effect* on the valuation function of a model. Transfer programs *only* affect the distribution of propositional variables to agents.

For the remaining constructs of transfer programs, we define the program relations inductively, in terms of the relations for atomic transfer programs, as defined above. Let the composition of relations $R_1$ and $R_2$ be denoted by $R_1 \circ R_2$, and the reflexive transitive closure (ancestral) of relation $R$ by $R^*$. Then the accessibility relations for complex programs are defined as follows (Harel et al., 2000):

$$\begin{array}{rcl} R_{\tau_1;\tau_2} & \hat{=} & R_{\tau_1} \circ R_{\tau_2} \\ R_{\tau_1 \cup \tau_2} & \hat{=} & R_{\tau_1} \cup R_{\tau_2} \\ R_{\tau^*} & \hat{=} & (R_\tau)^* \\ R_{\varphi?} & \hat{=} & \{(\mathcal{M}, \mathcal{M}) \mid \mathcal{M} \models^{\mathsf{d}} \varphi\}. \end{array}$$

Notice that the last of these definitions refers to the relation $\models^{\mathsf{d}}$, which of course has not yet been defined. The aim of the next section is to define this relation. We emphasise that, although the relations $R_{\varphi?}$ and $\models^{\mathsf{d}}$ mutually refer to one-another, both relations are in fact well-defined (as in conventional dynamic logic).

### 2.5 Truth Conditions

We interpret formulas of DCL-PC with respect to models, as introduced above. Given a model $\mathcal{M} = \langle \mathbb{A}, \mathbb{P}, \xi_0, \theta \rangle$ and a formula $\varphi$, we write $\mathcal{M} \models^{\mathsf{d}} \varphi$ to mean that $\varphi$ is satisfied (or, equivalently, true) in $\mathcal{M}$, under the 'direct' semantics. The rules defining the satisfaction relation $\models^{\mathsf{d}}$ are as follows:

$\mathcal{M} \models^{\mathsf{d}} \top$

$\mathcal{M} \models^{\mathsf{d}} p$ iff $\theta(p) = \mathrm{tt}$ (where $p \in \mathbb{P}$)

$\mathcal{M} \models^{\mathsf{d}} \neg\varphi$ iff $\mathcal{M} \not\models^{\mathsf{d}} \varphi$

$\mathcal{M} \models^{\mathsf{d}} \varphi \vee \psi$ iff $\mathcal{M} \models^{\mathsf{d}} \varphi$ or $\mathcal{M} \models^{\mathsf{d}} \psi$

$\mathcal{M} \models^{\mathsf{d}} \Diamond_C \varphi$ iff there exists a $C$-valuation $\theta_C$ such that $\mathcal{M} \oplus \theta_C \models^{\mathsf{d}} \varphi$

$\mathcal{M} \models^{\mathsf{d}} \langle\tau\rangle\varphi$ iff there exists a model $\mathcal{M}'$ such that $(\mathcal{M}, \mathcal{M}') \in R_\tau$ and $\mathcal{M}' \models^{\mathsf{d}} \varphi$.

We say a formula is *objective* if it contains no modal constructs (i.e., operators $\Diamond_C$ or $\langle\tau\rangle$). Thus objective formulae are formulae of classical propositional logic.

We assume the conventional definitions of satisfiability and validity: a DCL-PC formula $\varphi$ is $^{\mathsf{d}}$-*satisfiable* iff there exists a DCL-PC model $\mathcal{M}$ such that $\mathcal{M} \models^{\mathsf{d}} \varphi$, and $\varphi$ is $^{\mathsf{d}}$-*valid* iff for every DCL-PC model $\mathcal{M}$ we have $\mathcal{M} \models^{\mathsf{d}} \varphi$. We write $\models^{\mathsf{d}} \varphi$ to indicate that $\varphi$ is $^{\mathsf{d}}$-valid. A valid formula is also called a tautology. We say $\varphi$ is *feasible* if it is satisfiable but not valid. Finally, for any set of formulas $\Gamma$ and formula $\varphi$, we define $\Gamma \models^{\mathsf{d}} \varphi$ as $\forall \mathcal{M}(\forall \gamma \in \Gamma \mathcal{M} \models^{\mathsf{d}} \gamma \Rightarrow \mathcal{M} \models^{\mathsf{d}} \varphi$.

Let us define the box '$\Box_{\ldots}$' to be the dual of the cooperative ability modality '$\Diamond_{\ldots}$' as:

$$\Box_C \varphi \hat{=} \neg \Diamond_C \neg \varphi$$

and the '$[\cdot]_{\ldots}$' to be the dual of the transfer modality '$\langle\cdot\rangle_{\ldots}$' as:

$$[\tau]\varphi \mathrel{\hat{=}} \neg\langle\tau\rangle\neg\varphi.$$

Where $C$ is a coalition and $\varphi$ is a formula of DCL-PC, we write *controls*$(C,\varphi)$ to mean that $C$ can choose $\varphi$ to be either true or false:

$$\mathit{controls}(C,\varphi) \mathrel{\hat{=}} \Diamond_C\varphi \wedge \Diamond_C\neg\varphi \tag{1}$$

By using the *controls*$(\cdot,\cdot)$ construct, we can capture the distribution of propositional variables among the agents in a model.

**Lemma 1** Let $\mathcal{M} = \langle \mathbb{A}, \mathbb{P}, \xi_0, \theta\rangle$ be a model for DCL-PC, $i \in \mathbb{A}$ an agent, $C \subseteq \mathbb{A}$ a set of agents, and $p \in \mathbb{P}$ a propositional variable in $\mathcal{M}$. Then

1. (van der Hoek & Wooldridge, 2005b) $\mathcal{M} \models^{\mathsf{d}} \mathit{controls}(i,p)$ iff $p \in \mathbb{P}_i$;
2. $\mathcal{M} \models^{\mathsf{d}} \mathit{controls}(C,p)$ iff $p \in \mathbb{P}_C$.

**Remark 1** *Can we characterize the formulas under control of a coalition C? We have:*

$$\text{If } \varphi \text{ is feasible and objective, then } \models^{\mathsf{d}} \Big(\bigwedge_{p\in\mathbb{P}(\varphi)} \mathit{controls}(C,p)\Big) \rightarrow \mathit{controls}(C,\varphi) \tag{2}$$

*Observe that property (2) is not true for* arbitrary DCL-PC *formulas. To see this, take for example the formula $\langle i \rightsquigarrow_p j\rangle\top$, no matter whether we define $\mathbb{P}(\langle i \rightsquigarrow_p j\rangle\top)$ to be $\{p\}$ or $\emptyset$. We have $\models^{\mathsf{d}}$ $\neg$controls$(i, \langle i \rightsquigarrow_p j\rangle\top)$: independent of i owning p, exactly one of the two formulas $\langle i \rightsquigarrow_p j\rangle\top$ and $\neg\langle i \rightsquigarrow_p j\rangle\top$ is true. That is, $p \in \mathbb{P}_i$ in $\mathcal{M}$ iff $\mathcal{M} \models^{\mathsf{d}} \langle i \rightsquigarrow_p j\rangle\top$.*

*Also note that the $\leftarrow$-direction of the right hand side of (2) is not valid for objective $\varphi$: suppose $\mathcal{M} = \langle \mathbb{A}, \mathbb{P}, \xi_0, \theta\rangle$ such that $\theta(q) = \mathrm{tt}$, $p \in \mathbb{P}_i$, and $q \notin \mathbb{P}_i$. Then, we have $\mathcal{M} \models^{\mathsf{d}}$ controls$(i, p\wedge q) \wedge \neg($controls$(i,p) \wedge$ controls$(i,q))$: because q 'happens' to be true in $\mathcal{M}$, i controls the conjunction $p \wedge q$, but not each of its conjuncts.*

### 2.6 A Kripke Semantics

Although the direct semantics naturally captures the notions of propositional control and transfer of control, for some purposes – establishing completeness in particular, and relating it to the main stream of modal logic – it is more convenient to formulate the semantics for DCL-PC using conventional Kripke structures (Chellas, 1980; Blackburn, de Rijke, & Venema, 2001). The idea is that, given the (fixed, finite, non-empty) set $\mathbb{A}$ of agents and (fixed, finite, non-empty) set $\mathbb{P}$ of propositional variables, there will be a possible world for every possible allocation of the variables in $\mathbb{P}$ to the agents in $\mathbb{A}$ and every possible propositional valuation function over $\mathbb{P}$. Between those worlds, there are basically two 'orthogonal' accessibility relations (cf. Figure 4): a 'horizontal' and a 'vertical' one. First of all, we have a 'horizontal' relation $R_i$ for agent $i$ between two worlds $u$ and $v$ if agent $i$ is able, given the valuation $\theta_u$ in $u$, to turn it into the valuation $\theta_v$ as described by $v$, just by choosing appropriate values for her variables. Formally, $(u,v) \in R_i$ iff $\theta_u = \theta_v \pmod{\mathbb{P}_i}$. That is, $R_i$ is an equivalence relation. In what follows, we drop the symbolic distinction between worlds and valuations, i.e., we use $\theta$ for denoting a world and a valuation interchangeably. Notice that the ‘horizontal’ relation does not affect the allocation $\xi$: it remains unchanged. Let us therefore define our Kripke models to be $\mathfrak{M} = \langle \Theta, R_{i \in \mathbb{A}}, \xi \rangle$, where $\Theta$ is the set of all valuations $\theta$ over $\mathbb{P}$. It is important to realize that the sets $\mathbb{A}$ of agents and $\mathbb{P}$ of variables are fixed, but the allocations of variables to agents may vary. We denote the set of all such Kripke models by $\mathcal{K}(\mathbb{A}, \mathbb{P})$. We will call a pair $(\mathfrak{M}, \theta)$ a *pointed Kripke model*, and we will sometimes omit the brackets for such a pair.

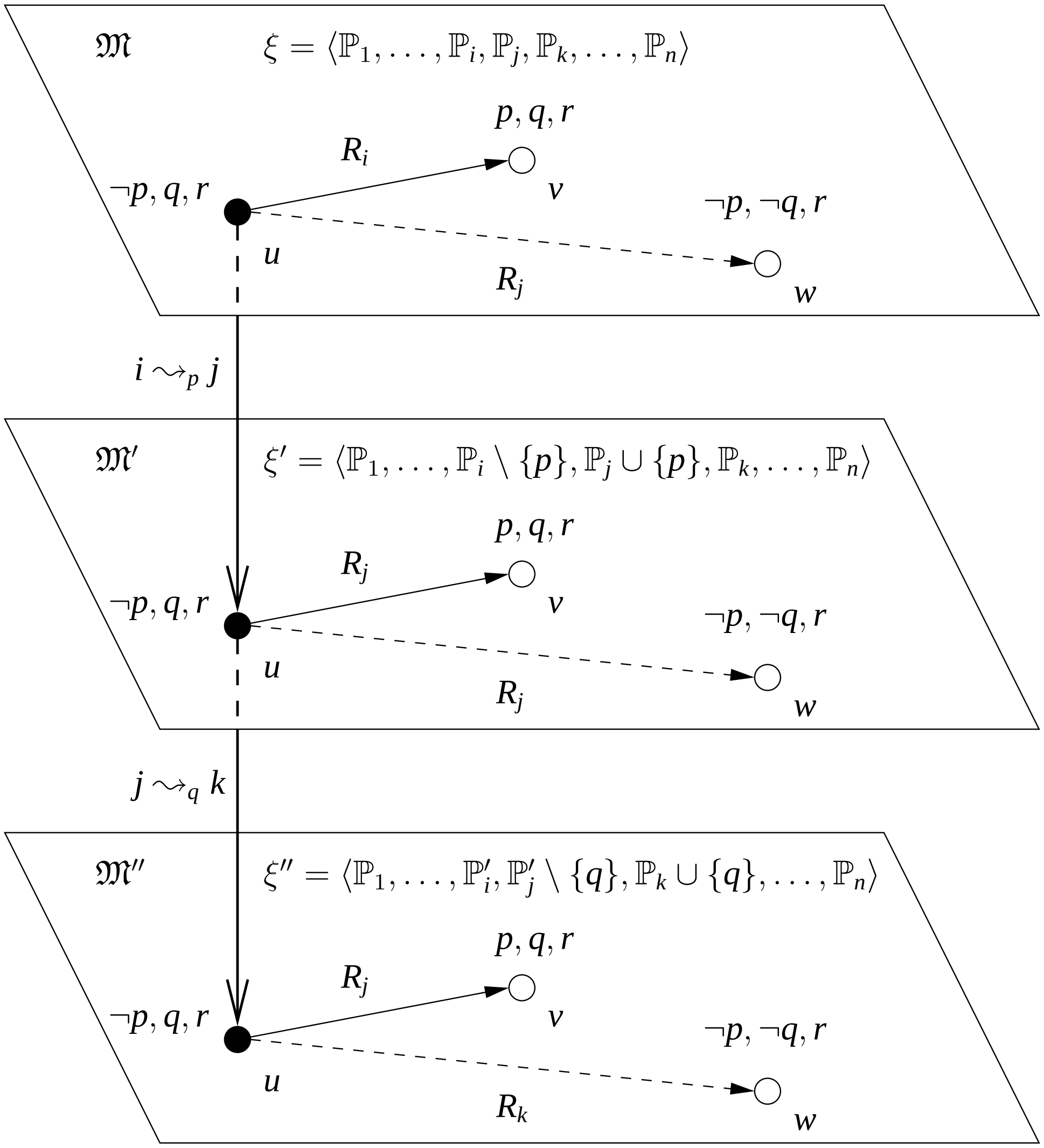

Figure 4: Some Kripke models for DCL-PC.

Secondly, the ‘vertical’ accessibility relation is between pointed models $(\mathfrak{M}, \theta)$ and $(\mathfrak{M}', \theta')$, where $\mathfrak{M} = \langle \Theta, R_{i \in \mathbb{A}}, \xi \rangle, \mathfrak{M}' = \langle \Theta, R_{i \in \mathbb{A}}, \xi' \rangle \in \mathcal{K}(\mathbb{A}, \mathbb{P})$, which indicate a change of the allocation $\xi$ to $\xi'$. Since such a change of allocation does not affect the current world, we have for such pairs that $\theta = \theta'$. Slightly abusing notation, we define $(\mathfrak{M}, \theta)(i \rightsquigarrow_p j)(\mathfrak{M}', \theta')$ exactly when $\theta = \theta'$ and $p \in \mathbb{P}_i$, and either $i = j$ and $\mathfrak{M} = \mathfrak{M}'$, or else $\mathbb{P}'_i = \mathbb{P}_i \setminus \{p\}$ and $\mathbb{P}'_j = \mathbb{P}_j \cup \{p\}$, and all the other sets $\mathbb{P}_h$ remain the same.

The truth relation $\models^{\mathsf{K}}$ interpreting formulas over Kripke structures holds between pairs of the form $(\mathfrak{M},\theta)$ and formulas $\varphi$. Its definition is as follows (we omit the Boolean cases and the cases for complex transfer programs):

> $\mathfrak{M},\theta \models^{\mathsf{K}} \Diamond_C\varphi$ iff there exists a valuation $\theta'$ such that $(\theta,\theta') \in R_i$ for each $i \in C$, and $\mathfrak{M},\theta' \models^{\mathsf{K}} \varphi$
>
> $\mathfrak{M},\theta \models^{\mathsf{K}} \langle i \rightsquigarrow_p j\rangle\varphi$ iff there exists a Kripke model $\mathfrak{M}'$ such that $(\mathfrak{M},\theta)(i \rightsquigarrow_p j)(\mathfrak{M}',\theta)$ and $\mathfrak{M}',\theta \models^{\mathsf{K}} \varphi$

For a set of formulas $\Gamma$ and a formula $\varphi$, we define $\Gamma \models^{\mathsf{K}} \varphi$ as $\forall(\mathfrak{M},\theta)(\forall\gamma \in \Gamma\ (\mathfrak{M},\theta) \models^{\mathsf{K}} \gamma \Rightarrow \mathfrak{M},\theta \models^{\mathsf{K}} \varphi)$. Figure 4 illustrates the Kripke semantics. Note that for the sets $\mathbb{P}'_i$ and $\mathbb{P}'_j$ in the Kripke model $\mathfrak{M}'$ we have $\mathbb{P}'_i = \mathbb{P}_i \setminus \{p\}$ and $\mathbb{P}'_j = \mathbb{P}_j \cup \{p\}$. Note that in the clause for $\Diamond_C\varphi$, the two pointed models $\mathfrak{M},\theta$ and $\mathfrak{M},\theta'$ are 'the same except for at most the atoms in $\mathbb{P}_C$'. This is a special case of two models being 'similar upto a set of atoms' (French, 2006; Ghilardi & Zawadowski, 2000).

**Remark 2** *Note that in fact, in the Kripke semantics, formulas are not interpreted in a model and a valuation only, but in the context of other models (which are reached by the atomic program $i \rightsquigarrow_p j$). There are finitely many of them, one for each $\xi$. Call this collection of models $\kappa$. In fact, this $\kappa$ is the structure with respect to which formulas are interpreted. In that sense, there is only one Kripke model for the language (w.r.t. $\mathbb{A},\mathbb{P}$): it is $\kappa$. We will prove completeness with respect to this unique two-dimensional model, in Section 3.*

The following lemma is easily established by induction on $\varphi$:

**Lemma 2** *For any fixed sets of agents $\mathbb{A}$ and propositional variables $\mathbb{P}$, the direct semantics and the Kripke semantics are equivalent, i.e., for any $\varphi$, any $\mathfrak{M} \in \mathcal{K}(\mathbb{A},\mathbb{P})$ with $\mathfrak{M} = \langle \Theta, R_{i\in\mathbb{A}}, \xi\rangle$, and any model $\mathcal{M} = \langle \mathbb{A},\mathbb{P},\xi,\theta\rangle$:*

$$\mathcal{M} \models^{\mathsf{d}} \varphi \ \textit{ iff } \ \mathfrak{M},\theta \models^{\mathsf{K}} \varphi.$$

As usual, we define $\mathfrak{M} \models^{\mathsf{K}} \varphi$ as $\forall\theta : \mathfrak{M},\theta \models^{\mathsf{K}} \varphi$, and $\models^{\mathsf{K}} \varphi$ as $\forall\mathfrak{M} : \mathfrak{M} \models^{\mathsf{K}} \varphi$.

## 3. A Complete Axiomatization

A sound and complete axiomatization for DCL-PC is presented in Figure 5. For the ease of exposition, we divide the axiomatization into five categories, as follows. While the 'Propositional Component' and the 'Rules of Inference' are straightforward, the 'Dynamic Component' is an immediate adaptation of Propositional Dynamic Logic (Harel et al., 2000). The 'Control Axioms' are inherited from CL-PC (van der Hoek & Wooldridge, 2005b). (The occurrence of $\ell(p)$ refers to a literal with atomic proposition $p$: it is either $p$ or $\neg p$, with the obvious meaning for $\neg\ell(p)$.) Note that *allocation* specifies that every propositional variable is assigned to exactly one agent (i.e., we have *an* allocation), while in contrast, for the fixed allocation $\xi$ that was assumed in CL-PC, one could explicitly state that $controls(i,p)$, for every $p \in \mathbb{P}_i$ (van der Hoek & Wooldridge, 2005b).

For the 'Transfer & Control Axioms', *atomic permanence* states that no program $\tau$ changes the valuation. From this, one easily extends this to arbitrary objective formulas (obtaining *objective permanence*, see Theorem 1 below). The axiom *persistence*$_1$(*control*) says that $i$'s control over $p$ is not affected when we move to another valuation, and axiom *persistence*$_2$(*control*) specifies how $i$ remains in control over $p$, even when a transfer program is executed: either the variable passed in that program is not $p$, or the delegating agent is not $i$. The axiom $precondition(transfer)$ expresses that agents can only give variables away that they possess, and, finally *func* says that the transition relation associated with an atomic transfer program is functional: at most one resulting world emerges.

| | | |
|---|---|---|
| <u>Propositional Component</u> | | |
| *Prop* | $\varphi$ | where $\varphi$ is any objective tautology |
| <u>Dynamic Component</u> | | |
| $K(\tau)$ | $[\tau](\varphi \rightarrow \psi) \rightarrow \big([\tau]\varphi \rightarrow [\tau]\psi\big)$ | |
| $union(\tau)$ | $[\tau \cup \tau']\varphi \leftrightarrow \big([\tau]\varphi \wedge [\tau']\varphi\big)$ | |
| $comp(\tau)$ | $[\tau;\tau']\varphi \leftrightarrow [\tau][\tau']\varphi$ | |
| $test(\tau)$ | $[\varphi?]\psi \leftrightarrow (\varphi \rightarrow \psi)$ | |
| $mix(\tau)$ | $\big(\varphi \wedge [\tau][\tau^*]\varphi\big) \leftrightarrow [\tau^*]\varphi$ | |
| $ind(\tau)$ | $\big(\varphi \wedge [\tau^*](\varphi \rightarrow [\tau]\varphi)\big) \rightarrow [\tau^*]\varphi$ | |
| <u>Control Axioms</u> | | |
| $K(i)$ | $\Box_i(\varphi \rightarrow \psi) \rightarrow \big(\Box_i\varphi \rightarrow \Box_i\psi\big)$ | |
| $T(i)$ | $\Box_i\varphi \rightarrow \varphi$ | |
| $B(i)$ | $\varphi \rightarrow \Box_i\Diamond_i\varphi$ | |
| *empty* | $\Box_\emptyset\varphi \leftrightarrow \varphi$ | |
| $control(i)$ | $controls(i,p) \leftrightarrow \big(\Diamond_i p \wedge \Diamond_i\neg p\big)$ | |
| *allocation* | $\bigwedge_{p\in\mathbb{P}} \big(controls(1,p) \triangledown \cdots \triangledown controls(n,p)\big)$ | where $\mathbb{A} = \{1,\ldots,n\}$ |
| $effect(i)$ | $\big(\psi \wedge \ell(p) \wedge controls(i,p)\big) \rightarrow \Diamond_i\big(\psi \wedge \neg\ell(p)\big)$ | where $\begin{cases} p \notin \mathbb{P}(\psi), \text{ and} \\ \psi \text{ is objective} \end{cases}$ |
| *Comp-*$\cup$ | $\Box_{C_1}\Box_{C_2}\varphi \leftrightarrow \Box_{C_1\cup C_2}\varphi$ | |
| <u>Transfer & Control Axioms</u> | | |
| *atomic permanence*$(\rightsquigarrow)$ | $\langle i \rightsquigarrow_p j\rangle\top \rightarrow \big([i \rightsquigarrow_p j]q \leftrightarrow q\big)$ | |
| $persistence_1(control)$ | $controls(i,p) \rightarrow \Box_j controls(i,p)$ | |
| $persistence_2(control)$ | $controls(i,p) \rightarrow [j \rightsquigarrow_q h]controls(i,p)$ | where $i \neq j$ or $p \neq q$ |
| $precondition(transfer)$ | $\langle i \rightsquigarrow_p j\rangle\top \rightarrow controls(i,p)$ | |
| *transfer* | $controls(i,p) \rightarrow \langle i \rightsquigarrow_p j\rangle controls(j,p)$ | |
| *func* | $controls(i,p) \rightarrow \big(\langle i \rightsquigarrow_p j\rangle\varphi \leftrightarrow [i \rightsquigarrow_p j]\varphi\big)$ | |
| <u>Rules of Inference</u> | | |
| *Modus Ponens* | $\vdash \varphi, \vdash (\varphi \rightarrow \psi) \;\Rightarrow \vdash \psi$ | |
| *Necessitation* | $\vdash \varphi \;\Rightarrow \vdash \Box\varphi$ | $\Box = [\tau], \Box_i$ |

Figure 5: Axiomatic System for DCL-PC.

The following theorem lists some properties of DCL-PC, where $controls(C,p)$ is defined in equation (1) above.

**Theorem 1**

1. *The axioms $K(i)$, $T(i)$, $B(i)$, and $effect(i)$ have coalitional counterparts $K(C)$, $T(C)$, $B(C)$, and $effect(C)$ that are all derivable for any coalition $C$.*

| |
|---|
| *at-least*(*control*) : <br> $\big(\ell(p) \wedge \mathit{controls}(i,p)\big) \rightarrow \Diamond_i \neg \ell(p)$ |
| *at-most*(*control*) : <br> $\ell(p) \rightarrow \big(\Diamond_i \neg \ell(p) \rightarrow \Box_j \ell(p)\big)$ $(i \neq j)$ |
| *non-effect*(*i*) : <br> $\big(\Diamond_i \ell(p) \wedge \neg \mathit{controls}(i,p)\big) \rightarrow \Box_i \ell(p)$ |
| *persistence*(*non-control*) : <br> $\neg \mathit{controls}(i,p) \leftrightarrow \Box_j \neg \mathit{controls}(i,p)$ |
| *objective permanence*($\leadsto$) : <br> $\langle i \leadsto_p j \rangle \top \rightarrow \big(\varphi \leftrightarrow [i \leadsto_p j]\varphi\big)$ where $\varphi$ is objective |
| *objective permanence* : <br> $\langle \tau \rangle \top \rightarrow \big(\varphi \leftrightarrow [\tau]\varphi\big)$ where $\varphi$ is objective |
| *inverse* : <br> $\mathit{controls}(i,p) \rightarrow \big(\varphi \leftrightarrow [i \leadsto_p j; j \leadsto_p i]\varphi\big)$ |
| *reverse* : <br> $[i \leadsto_p j][k \leadsto_q h]\varphi \leftrightarrow [k \leadsto_q h][i \leadsto_p j]\varphi$ <br> where ($j \neq k$ and $h \neq i$) or $p \neq q$ |

Figure 6: Some Theorems of DCL-PC.

2. *Moreover, we know (van der Hoek & Wooldridge, 2005b) that the axioms* $K(i)$*,* $T(i)$*,* $B(i)$*, and* $\mathit{effect}(i)$ *have the coalitional counterparts* $K(C)$*,* $T(C)$*,* $B(C)$*, and* $\mathit{effect}(C)$ *that are derivable for any coalition C.*

3. $\vdash \mathit{controls}(C,p) \leftrightarrow \bigvee_{i \in C} \mathit{controls}(i,p)$.

4. $\vdash \mathit{controls}(C,p) \rightarrow \Box_j \mathit{controls}(C,p)$*, i.e., the property* $\mathit{persistence}_1(\mathit{control})$ *is also derivable when we replace agent i by an arbitrary coalition C.*

**Proof:** See Appendix A. QED

Consider the language without dynamic transfer operators, in which we only have propositional logic with cooperation modalities $\Diamond_C$. Models for these are $\mathfrak{M}$, $\mathfrak{M}' \in \mathcal{K}(\mathbb{A}, \mathbb{P})$. In this program-free language, every formula is equivalent to one without any occurrences of coalition operators van der Hoek and Wooldridge (2005b). For instance, suppose that $\mathbb{P}_i = \{p, q\}$. Then a formula $\Diamond_i(\neg p \wedge r)$ is equivalent to $(p \wedge r) \vee (\neg p \wedge r)$ (we 'read off' the current value of variable $r$ outside $i$'s control).

We now establish a similar result for the language including programs. Any world $(\mathfrak{M}, \theta)$ is completely characterized when we know which variables are true in it, and what the allocation of variables to agents is. In such a case, the truth of all objective formulas, formulas involving abilities and transfer programs is completely determined.

**Lemma 3** *Let* $\varphi$ *be an arbitrary* DCL-PC *formula and* $\zeta$ *a conjunction of assertions of the form* $\mathit{controls}(j,p)$ *or* $\neg \mathit{controls}(j,p)$*. Then, in* DCL-PC*, we can derive*

$$\vdash \Diamond_C(\varphi \wedge \zeta) \leftrightarrow (\zeta \wedge \Diamond_C \varphi).$$

**Proof:** Since by *Comp*-$\cup$, for $C = \{a_1, a_2, \ldots, a_C\}$, we have $\Diamond_C\psi \leftrightarrow \Diamond_{a_1}\Diamond_{a_2}\cdots\Diamond_{a_C}\psi$, it is sufficient to prove the claim for an individual agent $i$. Moreover, we can move all conjuncts of $\zeta$ out one by one, once we know that

$$\vdash \Diamond_i(\varphi \wedge \pm \mathit{controls}(j,p)) \leftrightarrow (\pm \mathit{controls}(j,p) \wedge \Diamond_i\varphi),$$

where $\pm\mathit{controls}(j,p)$ is either $\mathit{controls}(j,p)$ or $\neg\mathit{controls}(j,p)$. We do the reasoning for the non-negated case (the other one is similar): $\Diamond_i(\varphi \wedge \mathit{controls}(j,p))$ is equivalent to

$$\begin{array}{c}\big(\mathit{controls}(j,p) \wedge \Diamond_i(\varphi \wedge \mathit{controls}(j,p))\big)\\ \vee\big(\neg\mathit{controls}(j,p) \wedge \Diamond_i(\varphi \wedge \mathit{controls}(j,p))\big).\end{array}$$

However, by using the theorem $\mathit{persistence}(\mathit{non\text{-}control})$ from Figure 6 (which we derive below), we have for the second disjunct that $\big(\neg\mathit{controls}(j,p) \wedge \Diamond_i(\varphi \wedge \mathit{controls}(j,p))\big) \leftrightarrow \bot$. That concludes the proof.

For $\mathit{persistence}(\mathit{non\text{-}control})$, the right-to-left direction follows immediately from $T(j)$. For the other direction, assume that $\neg\mathit{controls}(i,p)$. From *allocation* we derive that

$$\mathit{controls}(1,p)\triangledown\cdots\triangledown\mathit{controls}(i-1,p)\triangledown\mathit{controls}(i+1,p)\triangledown\cdots\triangledown\mathit{controls}(n,p),$$

and from this, by $\mathit{persistence}_1(\mathit{control})$, we get $\bigvee_{k\neq i}\Box_j\mathit{controls}(k,p)$. For every $k \neq i$, we have $\mathit{controls}(k,p) \rightarrow \neg\mathit{controls}(i,p)$, which follows from *allocation*. Hence, using *Necessitation*, we have $\Box_j(\mathit{controls}(k,p) \rightarrow \neg\mathit{controls}(i,p))$. From Axiom $K(j)$, it now follows that $\Box_j\mathit{controls}(k,p) \rightarrow \Box_j\neg\mathit{controls}(i,p)$. Combining this with $\bigvee_{k\neq i}\Box_j\mathit{controls}(k,p)$, we obtain the desired conclusion $\Box_j\neg\mathit{controls}(i,p)$. QED

Soundness of the axiom schemes in Figure 5 is readily checked. We now proceed to prove that the axiomatic system for DCL-PC in Figure 5 is complete. First, we introduce some notation.

**Definition 1** Given the set of propositional variables $\mathbb{P}$, a *valuation description* $\pi$ is a conjunction of literals ($p$ or $\neg p$) over them such that every propositional variable in $\mathbb{P}$ occurs in one literal.

Notice that, for each propositional variable $p \in \mathbb{P}$, it holds that either $\pi \rightarrow p$, or $\pi \rightarrow \neg p$. We denote the set of all valuation descriptions over $\mathbb{P}$ with $\Pi$. Notice that, for each valuation $\theta$, there is a $\pi_\theta \in \Pi$ such that

$$\begin{array}{rl}\pi_\theta = & \bigwedge\{p \mid p \in \mathbb{P} \text{ and } \theta(p) = \mathrm{tt}\}\\ & \wedge\bigwedge\{\neg p \mid p \in \mathbb{P} \text{ and } \theta(p) = \mathrm{ff}\}.\end{array}$$

**Definition 2** Given the set of propositional variables $\mathbb{P}$ and the set of agents $\mathbb{A}$, an *allocation description* $\upsilon$ is a conjunction of formulas of the form $\mathit{controls}(i,p)$ where for every $p \in \mathbb{P}$, there is exactly one $i \in \mathbb{A}$ such that $\mathit{controls}(i,p)$ appears in $\upsilon$.

We denote the set of all allocation descriptions with $\Upsilon$. Notice that allocations $\xi$ and conjunctions $\upsilon$ correspond to each other: For each allocation $\xi = \langle\mathbb{P}_1,\ldots,\mathbb{P}_n\rangle$ of the variables in $\mathbb{P}$ over the agents in $\mathbb{A}$, there is a $\upsilon_\xi \in \Upsilon$ such that

$$\upsilon_\xi = \bigwedge_{i\in\mathbb{A},p\in\mathbb{P}_i}\mathit{controls}(i,p).$$

Therefore, we refer to formulas $\upsilon$ as allocation descriptions. Given two allocation descriptions $\upsilon, \upsilon' \in \Upsilon$, we say $\upsilon(i \rightsquigarrow_p j)\upsilon'$ if the following three conditions are satisfied: $\upsilon \rightarrow \mathit{controls}(i, p)$, $\upsilon' \rightarrow \mathit{controls}(j, p)$, and $\upsilon$ and $\upsilon'$ agree on all other *control* expressions.

**Definition 3** Let, for any allocation description $\upsilon$, $\Pi_\upsilon \subseteq \Pi$ be a set of valuation descriptions. Then, a formula of the form

$$\bigvee_{\upsilon \in \Upsilon} \left( \bigvee \Pi_\upsilon \wedge \upsilon \right) \tag{3}$$

will be called a *proposition description*.

We will later, in Theorem 2, see that every formula $\varphi$ is equivalent to a proposition description. The intuition here is that the truth of $\varphi$ requires, for every allocation description $\upsilon$, some possible truth values of atoms to be fixed. To give an example, suppose there are two agents $i$ and $j$, and three atoms $p$, $q$ and $r$. Consider the formula $\varphi = \langle i \rightsquigarrow_p j \rangle (q \wedge \Diamond_j (p \wedge r))$. In order to find the equivalent proposition description, we must, for every $\upsilon \in \Upsilon$ make proper choices for $\Pi_\upsilon$. If $\upsilon$ implies $\neg \mathit{controls}(i, p)$ this allocation would make $\varphi$ false (since $i$ cannot transfer control over $p$), so for those $\upsilon$, we have to choose $\Pi_\upsilon$ to be the empty set, ensuring that $(\bigvee \Pi_\upsilon \wedge \upsilon)$ is equivalent to $\bot$. If $\upsilon$ implies $\mathit{controls}(i, p)$, there are basically two cases: either $\upsilon$ also implies $\mathit{controls}(j, r)$, and the constraint on $\Pi_\upsilon$ is that $\bigvee \Pi_\upsilon$ is equivalent to $q$, or else $\upsilon$ implies $\neg \mathit{controls}(j, r)$, in which case $\bigvee \Pi_\upsilon$ is equivalent to $q \wedge r$.

Let us, for two valuation descriptions $\pi$ and $\pi'$, a coalition $C$, and an allocation description $\upsilon$, write $\pi \equiv \pi' \pmod{C, \upsilon}$ if the two conjunctions of literals $\pi$ and $\pi'$ only differ in the variables under control of $C$, which is determined by $\upsilon$. For instance, when $C = \{1, 2\}$ and $\upsilon = \mathit{controls}(1, p_1) \wedge \mathit{controls}(2, p_2) \wedge \mathit{controls}(3, p_3)$, then $(\neg p_1 \wedge p_2 \wedge p_3) \equiv (p_1 \wedge \neg p_2 \wedge p_3) \pmod{C, \upsilon}$.

We now first collect some facts about *valuation descriptions*, *allocation descriptions*, and *proposition descriptions*. Recall that, for any set $\Phi$ of DCL-PC formulas, $\triangledown_{\varphi \in \Phi} \varphi$ is used as shorthand for $\bigvee_{\varphi \in \Phi} \varphi \wedge \bigwedge_{\varphi_1 \neq \varphi_2 \in \Phi} \neg(\varphi_1 \wedge \varphi_2)$.

**Lemma 4** *Given the set $\Pi$ of valuation descriptions $\pi$, and the set $\Upsilon$ of allocation descriptions $\upsilon$, the following six items are satisfied:*

1. $\vdash \triangledown_{\pi \in \Pi}\, \pi$
2. $\vdash \triangledown_{\upsilon \in \Upsilon}\, \upsilon$
3. $\vdash \varphi \rightarrow \bigvee_{\upsilon \in \Upsilon} (\varphi \wedge \upsilon)$
4. *For all $\upsilon \in \Upsilon$ and all $\pi \in \Pi$:* $\vdash \upsilon \rightarrow (\Diamond_C \pi \leftrightarrow \bigvee_{\pi' \equiv \pi \pmod{C, \upsilon}} \pi')$.
5. $\vdash (\upsilon \wedge \langle i \rightsquigarrow_p j \rangle \upsilon') \rightarrow ((\upsilon \wedge \pi) \leftrightarrow \langle i \rightsquigarrow_p j \rangle (\upsilon' \wedge \pi))$.
6. *Let $n$ be the number of agents, and $k$ the number of propositional variables. Then there are not more than $N(n, k) = 2^{2^{nk}}$ provably non-equivalent proposition descriptions.*

**Proof:**

1. This follows from *Prop* and the definition of $\Pi$: the $\pi$'s are mutually exclusive and cannot all be false.

2. Item (2) is easily seen to be equivalent to the *allocation* axiom: *allocation* implies $\bigvee_{\upsilon \in \Upsilon} \upsilon$, and, for every allocation description $\upsilon \in \Upsilon$, we have that $\upsilon$ implies *allocation*.

3. Item (3) is immediate from Item (2) and axiom *Prop*. In particular, using *Prop* we derive from $\vdash A \vee B$ that $C \rightarrow (C \wedge A) \vee (C \wedge B)$.

4. Assume $\upsilon$. For the right-to-left direction, also assume $\pi'$, for some valuation description $\pi'$ with $\pi' \equiv \pi \pmod{C, \upsilon}$. This means that $\pi'$ and $\pi$ only differ in some variables $p_1, \ldots, p_m$ for which $controls(C, p_j)$ is implied by $\upsilon$ (for $j = 1 \ldots m$). Note that $\pi'$ is an objective formula. We can write $\pi'$ as $\psi_i \wedge \ell(p_i)$ (for $i = 1 \ldots m$), where $\psi_i$ is as $\pi'$ but with the literal $\ell(p_i)$ left out. Apparently, we have $\psi_1 \wedge \ell(p_1) \wedge controls(C, p_1)$. Since $\psi_1$ is objective, we can apply $effect(C)$ to conclude $\Diamond_C(\psi_1 \wedge \neg\ell(p_1))$. Using Lemma 3, we derive $\Diamond_C(\upsilon \wedge \psi_1 \wedge \neg\ell(p_1))$. We can now rewrite $\psi_1$ to $\psi_2 \wedge \ell(p_2)$, and obtain $\Diamond_C(\psi_2 \wedge \ell(p_2) \wedge \neg\ell(p_1) \wedge \upsilon)$. We use $effect(C)$ and Lemma 3 again and get $\Diamond_C\Diamond_C(\psi_2 \wedge \neg\ell(p_2) \wedge \neg\ell(p_1) \wedge \upsilon)$. By *Comp*-$\cup$, this is the same as $\Diamond_C(\psi_2 \wedge \neg\ell(p_2) \wedge \neg\ell(p_1) \wedge \upsilon)$. We can repeat this process until we get $\Diamond_C(\psi_j \wedge \neg\ell(p_j) \wedge \cdots \wedge \neg\ell(p_2) \wedge \neg\ell(p_1) \wedge \upsilon)$. But, by definition of $\pi'$, this implies $\Diamond_C\pi$.

   We show the other direction from left to right by contrapositive: fix $\pi$ and $\upsilon$ and assume $\neg \bigvee_{\pi' \equiv \pi \pmod{C,\upsilon}} \pi'$. We have to show that $\neg\Diamond_C\pi$. Let $Q(\pi, \upsilon) = \{\ell(p) \mid \vdash \pi \rightarrow \ell(p)$ and $\nvdash \upsilon \rightarrow controls(C, p)\}$ be the set of the literals over variables that are not under control of the agents in $C$ at the allocation $\upsilon$. Notice that all valuations $\pi' \equiv \pi \pmod{C, \upsilon}$ agree with $\pi$ on the literals in $Q(\pi, \upsilon)$. We can use propositional reasoning to derive from $\neg \bigvee_{\pi' \equiv \pi \pmod{C,\upsilon}} \pi'$ that $\neg \bigwedge_{\ell(p) \in Q(\pi,\upsilon)} \ell(p)$. Using $T(C)$ and $K(C)$, we then conclude that $\bigvee_{\ell(p) \in Q(\pi,\upsilon)} \Diamond_C \neg\ell(p)$. From $\upsilon$, it follows, for each literal $\ell(p) \in Q(\pi, \upsilon)$, that $\neg controls(C, \ell(p))$, which, by equation (1), equals $\Diamond_C\neg\ell(p) \rightarrow \neg\Diamond_C\ell(p)$. But then, we can derive $\bigvee_{\ell(p) \in Q(\pi,\upsilon)} \neg\Diamond_C\ell(p)$. Using $K(C)$, we obtain $\neg\Diamond_C \bigwedge_{\ell(p) \in Q(\pi,\upsilon)} \ell(p)$. Hence, $\neg\Diamond_C\pi$.

5. First of all, from $(\upsilon \wedge \langle i \rightsquigarrow_p j\rangle\upsilon'$, follows $\upsilon \rightarrow controls(i, p)$. Hence, given $\upsilon$, we have that all formulas $\langle i \rightsquigarrow_p j\rangle\psi$ and $[i \rightsquigarrow_p j]\psi$ are equivalent by *func*. In particular, we have $\langle i \rightsquigarrow_p j\rangle\top$. Let us first show for literals $\ell(q)$ that $\ell(q) \leftrightarrow \langle i \rightsquigarrow_p j\rangle\ell(q)$. For negative literals $\neg q$, $\neg q \leftrightarrow \langle i \rightsquigarrow_p j\rangle\neg q$ equals $q \leftrightarrow [i \rightsquigarrow_p j]q$, which follows from *atomic permanence*$(\rightsquigarrow)$. For positive literals $q$, we use *func* to obtain $q \leftrightarrow [i \rightsquigarrow_p j]q$, which again holds by *atomic permanence*$(\rightsquigarrow)$. Now, given $\upsilon$, we have $q \leftrightarrow \langle i \rightsquigarrow_p j\rangle q$ and $\neg q \leftrightarrow \langle i \rightsquigarrow_p j\rangle\neg q$. Then, for any valuation description $\pi$, we also have $\pi \leftrightarrow \langle i \rightsquigarrow_p j\rangle\pi$. It remains to show that $\upsilon \leftrightarrow \langle i \rightsquigarrow_p j\rangle\upsilon'$. From left to right, note that we have $controls(i, p) \rightarrow \langle i \rightsquigarrow_p j\rangle controls(j, p)$ by *transfer*. For any other *controls* expression implied by $\upsilon$, this follows from $persistence_2(control)$ and *func*. Finally, consider the direction from right to left: $\langle i \rightsquigarrow_p j\rangle\upsilon' \rightarrow \upsilon$. We have to show that, first, $\langle i \rightsquigarrow_p j\rangle controls(j, p) \rightarrow controls(i, p)$, and, second, that $\langle i \rightsquigarrow_p j\rangle controls(h, q) \rightarrow controls(h, q)$, for each $controls(h, q)$ (with $q \neq p$) implied by $\upsilon'$. The first part follows immediately from $precondition(transfer)$. For the second part, let $h$ be an agent and $q \neq p$ a variable such that $\upsilon'$ implies $controls(h, q)$. Suppose $\langle i \rightsquigarrow_p j\rangle controls(h, q)$. By *allocation*, we have that $\langle i \rightsquigarrow_p j\rangle \bigwedge_{k \neq h} \neg controls(k, q)$. But then the contrapositive of $persistence_2(control)$ yields $\bigwedge_{k \neq h} \neg controls(k, q)$. It follows by *allocation* that $controls(h, q)$.

6. Item 6 follows from a straightforward counting argument of proposition description formulas. The number of proposition descriptions depends on the cardinalities of the sets $\Pi$ and $\Upsilon$. Given the number $n = |\mathbb{A}|$ of agents, and the number $k = |\mathbb{P}|$ of propositional variables, it is easy to see that there are $2^k$ valuation descriptions in $\Pi$, and $n^k$ allocation descriptions in $\Upsilon$ (i.e., the number of ways we can distribute $k$ variables over $n$ agents). Observe that any proposition description formula is obtained by assigning a set of valuation descriptions $\pi$ to each allocation description $\upsilon$. Hence, there are $2^{2^k \times n^k}$ proposition descriptions. Since $2^{2^k \times n^k} \leq 2^{2^{nk}}$, we obtain with $N(n,k) = 2^{2^{nk}}$ an upper bound for the number of different proposition description formulas.

QED

We now present the main result of this section. We first formulate it, reflect briefly on it, and then give its proof.

**Theorem 2** *For every* DCL-PC *formula* $\varphi$, *there are sets* $\Pi_\upsilon(\varphi) \subseteq \Pi$ *of valuation descriptions, one for each* $\upsilon \in \Upsilon$, *such that*

$$\vdash \varphi \leftrightarrow \bigvee_{\upsilon \in \Upsilon} \big( \bigvee \Pi_\upsilon(\varphi) \wedge \upsilon \big).$$

According to Theorem 2, we can get rid of all the $\langle i \rightsquigarrow_p j \rangle \psi$. To show that our derivation system is good enough to establish that, is the main task of its proof. But let us first convince ourselves semantically that such a normal form makes sense. Remember that every model $\mathfrak{M}$ comes with its own allocation $\xi$. A formula like $\langle i \rightsquigarrow_p j \rangle \psi$ is true at $\mathfrak{M}, \theta$, if $\psi$ is true in a model that looks like $\mathfrak{M}$, but in which control over $p$ is transferred from $i$ to $j$. But this means that some formula $\psi'$ must already be true at $\mathfrak{M}, \theta$, where $i$ takes the role of $j$ for as far as $p$ is concerned. For instance, $\langle i \rightsquigarrow_p j \rangle (q \wedge \Diamond_j (p \wedge r))$ is true at $\mathfrak{M}, \theta$, if $q \wedge \Diamond_j r \wedge controls(i,p)$ is true under the current allocation $\xi$. This formula has no reference to other 'layers' anymore. More precisely, $\langle i \rightsquigarrow_p j \rangle (q \wedge \Diamond_j (p \wedge r))$ is equivalent to

$$\big(q \wedge ((p \wedge r) \vee (\neg p \wedge r)) \wedge controls(i,p) \wedge \neg controls(j,r)\big)$$
$$\vee$$
$$\big(q \wedge ((p \wedge r) \vee (\neg p \wedge r) \vee (p \wedge \neg r) \vee (\neg p \wedge \neg r)) \wedge controls(i,p) \wedge controls(j,r)\big).$$

**Proof:** The proof is by induction on the norm $||\cdot||$ which is defined on DCL-PC formulas as follows:

$$\begin{array}{rcl} ||\top|| &=& ||p|| = 0, \text{ for any } p \in \mathbb{P} \\ ||\neg\psi|| &=& 1 + ||\psi|| \\ ||\psi_1 \vee \psi_2|| &=& 1 + ||\psi_1|| + ||\psi_2|| \\ ||\Diamond_C \psi|| &=& 1 + ||\psi||, \text{ for any } C \subseteq \mathbb{A} \\ ||[i \rightsquigarrow_p j]\psi|| &=& 1 + ||\psi|| \\ ||[\varphi?]\psi|| &=& 1 + ||\neg\varphi \vee \psi|| \\ ||[\tau_1 \cup \tau_2]\psi|| &=& 1 + ||[\tau_1]\psi \vee [\tau_2]\psi|| \\ ||[\tau_1;\tau_2]\psi|| &=& 1 + ||[\tau_1][\tau_2]\psi|| \\ ||[\tau^*]\psi|| &=& 1 + ||\bigwedge_{i=0..N} [\tau]^i \psi|| \end{array}$$

where $N = N(n,k)$ is the number defined in Lemma 4, Item (6).

The induction base for the proof of our theorem has two cases:

- $\varphi = \top$. We take $\Pi_\upsilon(\top) = \Pi$, for every $\upsilon \in \Upsilon$. By Item (1) of Lemma 4, we have that $\bigvee \Pi$ is an objective tautology. Hence, $\bigvee_{\upsilon \in \Upsilon} \big( \bigvee \Pi_\upsilon(\top) \wedge \upsilon \big)$ is equivalent to $\bigvee_{\upsilon \in \Upsilon} \upsilon$, which in turn is equivalent to $\top$ by (2) of Lemma 4.

- $\varphi = p$, for $p \in \mathbb{P}$. Take $\Pi_\upsilon(p) = \{\pi \in \Pi \mid \vdash \pi \to p\}$, for every $\upsilon \in \Upsilon$. Clearly, $p$ is equivalent to $\bigvee \Pi_\upsilon(p)$, i.e., $\vdash p \leftrightarrow \bigvee \Pi_\upsilon(p)$. Now, using (3) of Lemma 4, we get $\vdash p \to \bigvee_{\upsilon\in\Upsilon}(p \wedge \upsilon)$. Finally, replacing the second occurrence of $p$ with its just derived equivalent formula, we get $\vdash p \to \bigvee_{\upsilon\in\Upsilon}\big(\bigvee \Pi_\upsilon(p) \wedge \upsilon\big)$. The other direction follows by simple propositional reasoning.

Consider the induction step.

- $\varphi = \neg\psi$. We set $\Pi_\upsilon(\neg\psi) = \Pi \setminus \Pi_\upsilon(\psi)$, for every $\upsilon \in \Upsilon$. This works, because of the following:

$$\neg\psi \leftrightarrow \neg \bigvee_{\upsilon\in\Upsilon} \big(\bigvee \Pi_\upsilon(\psi) \wedge \upsilon\big) \tag{4}$$

$$\leftrightarrow \bigwedge_{\upsilon\in\Upsilon} \big((\neg \bigvee \Pi_\upsilon(\psi)) \vee \neg\upsilon\big) \tag{5}$$

$$\leftrightarrow \bigvee_{\upsilon\in\Upsilon} \big((\neg \bigvee \Pi_\upsilon(\psi)) \wedge \upsilon\big) \tag{6}$$

$$\leftrightarrow \bigvee_{\upsilon\in\Upsilon} \big(\bigvee \Pi_\upsilon(\neg\psi) \wedge \upsilon\big) \tag{7}$$

  All steps are purely propositional, except for the equivalence between (5) and (6) which we explain now. Abbreviate (5) to $\bigwedge_{\upsilon\in\Upsilon}(\neg A_\upsilon \vee \neg\upsilon)$, then (6) is $\bigvee_{\upsilon\in\Upsilon}(\neg A_\upsilon \wedge \upsilon)$. Note that by Lemma 4, Item (2), we derive $\bigvee_{\upsilon\in\Upsilon} \upsilon$. In other words, one $\upsilon$ must be true, say $\hat\upsilon$. But note that $\bigwedge_{\upsilon\in\Upsilon}(\neg A_\upsilon \vee \neg\upsilon) \wedge \hat\upsilon$ implies $(\neg A_{\hat\upsilon} \wedge \hat\upsilon)$ and hence also $\bigvee_{\upsilon\in\Upsilon}(\neg A_\upsilon \wedge \upsilon)$, our abbreviation for (6). Conversely, if $\bigvee_{\upsilon\in\Upsilon}(\neg A_\upsilon \wedge \upsilon)$ holds, we know by Lemma 4, item (2) that $\triangle_{\upsilon\in\Upsilon}\upsilon$, i.e., exactly one $\upsilon$ must hold, say it is $\hat\upsilon$. For this $\hat\upsilon$ the formula $(\neg A_{\hat\upsilon} \wedge \hat\upsilon)$ is true, and it is the only allocation description for which such a formula can be true. For this $\hat\upsilon$, we also have $(\neg A_{\hat\upsilon} \vee \neg\hat\upsilon)$. Moreover, for any $\upsilon \neq \hat\upsilon$, we have $\neg\upsilon$, and hence $(\neg A_\upsilon \vee \neg\upsilon)$. So $(\neg A_\upsilon \vee \neg\upsilon)$ holds for all $\upsilon$, hence we have $\bigwedge_{\upsilon\in\Upsilon}(\neg A_\upsilon \vee \neg\upsilon)$, our shorthand for (5).

- $\varphi = \psi_1 \vee \psi_2$. We set $\Pi_\upsilon(\psi_1 \vee \psi_2) = \Pi_\upsilon(\psi_1) \cup \Pi_\upsilon(\psi_2)$, for every $\upsilon \in \Upsilon$. For the following equivalences, we only need propositional reasoning:

$$\psi_1 \vee \psi_2 \leftrightarrow \bigvee_{\upsilon\in\Upsilon} \big(\bigvee \Pi_\upsilon(\psi_1) \wedge \upsilon\big) \vee \bigvee_{\upsilon\in\Upsilon} \big(\bigvee \Pi_\upsilon(\psi_2) \wedge \upsilon\big) \tag{8}$$

$$\bigvee_{\upsilon\in\Upsilon} \big((\bigvee \Pi_\upsilon(\psi_1) \vee \bigvee \Pi_\upsilon(\psi_2)) \wedge \upsilon\big) \tag{9}$$

$$\bigvee_{\upsilon\in\Upsilon} \big(\bigvee \Pi_\upsilon(\psi_1 \vee \psi_2) \wedge \upsilon\big) \tag{10}$$

- $\varphi = \Diamond_C\psi$. For every $\upsilon \in \Upsilon$, we set

$$\Pi_\upsilon(\Diamond_C\psi) = \{\pi \in \Pi \mid \pi \equiv \pi' \pmod{C, \upsilon} \text{ for some } \pi' \in \Pi_\upsilon(\psi)\}$$

We can derive the following equivalences:

$$\begin{aligned}
\Diamond_C\psi &\leftrightarrow \Diamond_C \bigvee_{\upsilon\in\Upsilon} \big(\bigvee \Pi_\upsilon(\psi) \wedge \upsilon\big) && (11)\\
&\leftrightarrow \bigvee_{\upsilon\in\Upsilon} \Diamond_C\big(\bigvee \Pi_\upsilon(\psi) \wedge \upsilon\big) && (12)\\
&\leftrightarrow \bigvee_{\upsilon\in\Upsilon} \big[\Diamond_C\big(\bigvee \Pi_\upsilon(\psi)\big) \wedge \upsilon\big] && (13)\\
&\leftrightarrow \bigvee_{\upsilon\in\Upsilon} \big(\bigvee \Pi_\upsilon(\Diamond_C\psi) \wedge \upsilon\big) && (14)
\end{aligned}$$

The equivalence in (11) holds by the induction hypothesis. Using $K(i)$, this is equivalent to (12) (for any diamond we have $\Diamond_C(\varphi \vee \psi) \leftrightarrow (\Diamond_C\varphi \vee \Diamond_C\psi)$). The equivalence of the latter and (13) is by Lemma 3.

It remains to show the equivalence of (13) and (14). We have $\Diamond_C \bigvee \Pi_\upsilon(\psi) = \Diamond_C \bigvee_{\pi\in\Pi_\upsilon(\psi)} \pi$. By $K(C)$ and *Comp*-$\cup$, this formula is equivalent to $\bigvee_{\pi\in\Pi_\upsilon(\psi)} \Diamond_C\pi$. Using Item (4) in Lemma 4, we see that this is equivalent to $\bigvee_{\pi\in\Pi_\upsilon(\psi)} \bigvee_{\pi'\mid\pi'\equiv\pi \pmod{C,\upsilon}} \pi'$. But this equals $\bigvee \Pi_\upsilon(\Diamond_C\psi)$ by definition of $\Pi_\upsilon(\Diamond_C\psi)$.

- $\varphi = [i \rightsquigarrow_p j]\psi$. We define $\Pi_\upsilon([i \rightsquigarrow_p j]\psi)$ as follows: for every $\upsilon \in \Upsilon$,

$$\Pi_\upsilon([i \rightsquigarrow_p j]\psi) = \begin{cases} \Pi_{\upsilon'}(\psi) & \text{if } \upsilon \rightarrow \textit{controls}(i,p) \\ & \text{where } \upsilon(i \rightsquigarrow_p j)\upsilon' \\ \Pi & \text{otherwise} \end{cases}$$

To see that this yields a formula of the right form that is equivalent to $[i \rightsquigarrow_p j]\psi$, let us first partition $\Upsilon$ in $\Upsilon^+(i,p) = \{\upsilon \in \Upsilon \mid \vdash \upsilon \rightarrow \textit{controls}(i,p)\}$ and $\Upsilon^-(i,p) = \{\upsilon \in \Upsilon \mid \vdash \upsilon \rightarrow \neg\textit{controls}(i,p)\}$. Now consider the following derivable equivalences:

$$\begin{aligned}
[i \rightsquigarrow_p j]\psi &\leftrightarrow [i \rightsquigarrow_p j] \bigvee_{\upsilon'\in\Upsilon} \big(\bigvee \Pi_{\upsilon'}(\psi) \wedge \upsilon'\big) && (15)\\
&\leftrightarrow \neg\textit{controls}(i,p) \\
&\qquad \vee(\textit{controls}(i,p) \wedge \langle i \rightsquigarrow_p j\rangle \bigvee_{\upsilon_0\in\Upsilon} \big(\bigvee \Pi_{\upsilon_0}(\psi) \wedge \upsilon_0\big)) && (16)\\
&\leftrightarrow \bigvee_{\upsilon\in\Upsilon^-(i,p)} \big(\bigvee \Pi \wedge \upsilon\big) \vee \\
&\qquad \bigvee_{\upsilon\in\Upsilon^+(i,p)} \big(\upsilon \wedge \langle i \rightsquigarrow_p j\rangle \bigvee_{\upsilon_0\in\Upsilon} \big(\bigvee \Pi_{\upsilon_0}(\psi) \wedge \upsilon_0\big)\big) && (17)\\
&\leftrightarrow \bigvee_{\upsilon\in\Upsilon} \big(\bigvee \Pi_\upsilon([i \rightsquigarrow_p j]\psi) \wedge \upsilon\big) && (18)
\end{aligned}$$

The equation in (15) holds by the induction hypothesis. It is equivalent to (16) by propositional reasoning, and changing from $[i \rightsquigarrow_p j]$ to $\langle i \rightsquigarrow_p j\rangle$ is allowed by *func*. The equivalence of (16) and (17) follows from the definition of $\Upsilon^+(i,p)$ and $\Upsilon^-(i,p)$ and the fact that $\bigvee \Pi \leftrightarrow \top$. In order to prove the equivalence between (17) and (18), it is sufficient to show that, for any fixed $\upsilon \in \Upsilon^+(i,p)$, the formula $\upsilon \wedge \langle i \rightsquigarrow_p j\rangle \bigvee_{\upsilon_0 \in \Upsilon}(\bigvee \Pi_{\upsilon_0}(\psi) \wedge \upsilon_0)$ is equivalent to $\upsilon \wedge \bigvee \Pi_{\upsilon'}(\psi)$, where $\upsilon(i \rightsquigarrow_p j)\upsilon'$. But this follows from Lemma 4, Item (5), as follows. First of all, we can write $\upsilon \wedge \langle i \rightsquigarrow_p j\rangle \bigvee_{\upsilon_0 \in \Upsilon}(\bigvee \Pi_{\upsilon_0}(\psi) \wedge \upsilon_0)$ as $\upsilon \wedge \bigvee_{\upsilon_0 \in \Upsilon} \langle i \rightsquigarrow_p j\rangle(\bigvee \Pi_{\upsilon_0}(\psi) \wedge \upsilon_0)$. By the mentioned lemma, we know exactly which $\upsilon_0$ we need: it is $\upsilon'$ for which $\upsilon(i \rightsquigarrow_p j)\upsilon'$ giving $\upsilon \wedge \langle i \rightsquigarrow_p j\rangle(\bigvee \Pi_{\upsilon'}(\psi) \wedge \upsilon')$. We can rewrite this into $\upsilon \wedge \langle i \rightsquigarrow_p j\rangle \bigvee\{\pi \wedge \upsilon' \mid \pi \in \Pi_{\upsilon'}(\psi)\}$, and then push the diamond $\langle i \rightsquigarrow_p j\rangle$ inside the disjunction to get $\upsilon \wedge \bigvee\{\langle i \rightsquigarrow_p j\rangle(\pi \wedge \upsilon') \mid \pi \in \Pi_{\upsilon'}(\psi)\}$. But then Lemma 4, Item (5) yields $\upsilon \wedge \bigvee \Pi_{\upsilon'}(\psi)$. The other direction is similar: if $\upsilon \wedge \bigvee \Pi_{\upsilon'}(\psi)$ and $\upsilon(i \rightsquigarrow_p j)\upsilon'$, then, by Lemma 4, Item (5), we get $\upsilon \wedge \langle i \rightsquigarrow_p j\rangle(\bigvee \Pi_{\upsilon'}(\psi) \wedge \upsilon')$, from which the result follows.

- $\varphi = [\psi'?]\psi$. By axiom *test*$(\tau)$, $[\psi'?]\psi$ is equivalent to $\neg\psi' \vee \psi$, which has an equivalent formula of the right form by the induction hypothesis.

- $\varphi = [\tau_1;\tau_2]\psi$. By axiom *comp*$(\tau)$, $[\tau_1;\tau_2]\psi$ is equivalent to $[\tau_1][\tau_2]\psi$, which has an equivalent formula of the right form by the induction hypothesis.

- $\varphi = [\tau_1 \cup \tau_2]\psi$. By axiom *union*$(\tau)$, $[\tau_1 \cup \tau_2]\psi$ is equivalent to $[\tau_1]\psi \vee [\tau_2]\psi$, which has an equivalent formula of the right form by the induction hypothesis.

- $\varphi = [\tau^*]\psi$. Recall $N = N(n,k)$ as given in Lemma 4, Item (6). Using axiom *mix*$(\tau)$ and $K(\tau)$, we know that $[\tau^*]\psi$ is equivalent to $\psi \wedge [\tau][\tau^*]\psi$. Doing this $N$ times, we obtain

$$\psi \wedge [\tau]\psi \wedge [\tau]^2\psi \wedge \cdots \wedge [\tau]^N\psi \wedge [\tau]^N[\tau^*]\psi.$$

By the induction hypothesis, we know that all except the last conjunct have an equivalent normal form. But since there are only $N$ different such forms, the conjunct $[\tau]^N\psi$ must be equivalent to one of the earlier conjuncts $[\tau]^i\psi$ ($i < N$). Define $\lambda = \bigwedge_{i=0..N}[\tau]^i\psi$. We claim

$$\lambda \leftrightarrow [\tau^*]\psi. \tag{19}$$

And by the induction hypothesis, $\lambda$ has an equivalent formula of the right form. The '$\leftarrow$' direction of (19) is obvious since $\lambda$ 'is just the first part' of the 'unraveling' $\psi\wedge[\tau]\psi\wedge[\tau][\tau]\psi\ldots$ of $[\tau^*]\psi$ using axioms *mix*$(\tau)$ and $K(\tau)$. To show the other direction it is sufficient to derive $\lambda \rightarrow [\tau^*]\lambda$, because, by the fact that $\psi$ is just one of the conjuncts in $\lambda$, this immediately gives $\lambda \rightarrow [\tau^*]\psi$. To show the derivability of $\lambda \rightarrow [\tau^*]\lambda$, we will use *ind*$(\tau)$. First of all, we show $\lambda \rightarrow [\tau]\lambda$. To see this, note that $[\tau]\lambda \leftrightarrow [\tau][\tau]^0\psi \wedge [\tau][\tau]^1\psi \wedge \cdots \wedge [\tau][\tau]^N\psi$. By the induction hypothesis, each conjunct $[\tau]^i\psi$ ($i \le N$) has a normal form, say, $B_i$. Then we obtain from $\lambda$ a sequence $B_0, B_1, \ldots, B_N$ of $N+1$ formulas in normal form. Since there are at most $N$ provably non-equivalent formulas by Lemma 4, Item (6), we know that there is a $B_j$ that equals a previous $B_a$ with $a < j \le N$. Let $B_j$ be the first such repetition in this sequence. Notice that we have now $[\tau]^j\psi \equiv B_j = B_a \equiv [\tau]^a\psi$, and thus $[\tau]^k[\tau]^j\psi \equiv [\tau]^k[\tau]^a\psi$, for any $k \ge 0$. But then, it follows that the last conjunct $[\tau][\tau]^N\psi$ in $[\tau]\lambda$ is equivalent to $[\tau]^k[\tau]^a\psi$ (with $k = N - j$), which already appears in $\lambda$. Now that we have derived $\lambda \rightarrow [\tau]\lambda$, we apply *Necessitation* for $[\tau^*]$, and obtain $\vdash [\tau^*](\lambda \rightarrow [\tau]\lambda)$. By applying *ind*$(\tau)$, we get $\lambda \rightarrow [\tau^*]\lambda$.

QED

We know from Lemma 4 that there are only finitely many different normal forms: since every formula has such a normal form, there can be only finitely many non-equivalent formulas. We also know from the proof of Theorem 2 above that, for $[\tau^*]\psi$, we only have to consider a finite number of conjuncts $[\tau]^i\psi$ in its unraveling.

**Corollary 1** *There are only finitely many pairwise non-equivalent formulas of* DCL-PC*. In fact, given the number n of agents, and k of propositional variables, we have:*

1. $\forall i \neq j \leq M \ \nvdash (\psi_i \leftrightarrow \psi_j) \ \Rightarrow \vdash \bigvee_{i \leq M} \psi_i$, *and*
2. $\vdash [\tau^*]\psi \leftrightarrow \bigwedge_{i \leq N} [\tau]^i\psi$,

*where* $M = 2^{nk}$ *and* $N = 2^{2^{nk}}$ *(as defined in Lemma 4, Item* (6)*).*

*Completeness* of a derivation system with inference relation $\vdash$ with respect to a semantics means that every semantically valid formula is also provable: $\models \varphi \ \Rightarrow \vdash \varphi$. In order to prove completeness, often the contrapositive of this is shown: $\nvdash \varphi \ \Rightarrow \not\models \varphi$. That is, every consistent formula $\neg\varphi$ has a model. A popular technique in modal logic is to construct a dedicated model – the canonical model, (cf. e.g., Blackburn et al., 2001) – for any consistent formula $\psi$. That canonical model is a 'bridge' between syntax and semantics: it consists of all maximal consistent sets $\Sigma$ (as worlds), and is constructed in such a way that membership of a world in $\Sigma$ and truth in the corresponding world coincide.

For DCL-PC this works straightforwardly as follows. Fix the finite sets $\mathbb{A}$ and $\mathbb{P}$, and take a consistent formula $\psi$. Build a maximal consistent set $\Delta$ 'around it'. Let the normal form of $\psi$, guaranteed by Theorem 2, be $\bigvee_{\upsilon}(\bigvee \Pi_{\upsilon} \wedge \upsilon)$. Since $\Delta$ is maximal consistent, for some $\upsilon$, we have $(\bigvee \Pi_{\upsilon} \wedge \upsilon) \in \Delta$. Again, by maximal consistency of $\Delta$, it must contain, for exactly one $\pi \in \Pi_{\upsilon}$, the formula $\pi \wedge \upsilon$. But $\pi$ uniquely determines a valuation $\theta = \theta_{\pi}$, whereas $\upsilon$ determines an allocation $\xi = \xi_{\upsilon}$. In other words, $\Delta$ uniquely determines a pointed Kripke model $(\mathfrak{M}, \theta)$ with $\mathfrak{M} = \langle \Theta, R_{i\in\mathbb{A}}, \xi \rangle$. All the worlds in $\Theta$ are determined by $\mathbb{P}$, and all $R_{i\in\mathbb{A}}$ (the horizontal layer of $\mathfrak{M}$, in terms of Figure 4) are determined by the Control Axioms. The availability of all the right models $\mathfrak{M}' = \langle \Theta, R_{i\in\mathbb{A}}, \xi' \rangle$ (the vertical layer in Figure 4) is determined by the Delegation and Control Axioms. As a result, we can directly interpret all subformulas of the form $\langle i \leadsto_p j \rangle\varphi$ and $\Diamond_C\varphi$ in the proper way, in $(\mathfrak{M}, \theta)$.

This argument easily extends to strong completeness, which states that for all sets of formulas $\Gamma$ and all formulas $\varphi$, we have $\Gamma \models^{\mathsf{K}} \varphi \Rightarrow \Gamma \vdash \varphi$. To see this, suppose $\Gamma \nvdash \varphi$, i.e., $\Gamma \cup \{\neg\varphi\}$ is consistent. Since there are only finitely many pairwise non-equivalent formulas, there must be a formula $\psi$ that is equivalent with $\Gamma \cup \{\neg\varphi\}$. By the previous argument, we find a pointed model $(\mathfrak{M}, \theta)$ such that $(\mathfrak{M}, \theta) \models^{\mathsf{K}} \psi$. For this model we also have $(\mathfrak{M}, \theta) \models^{\mathsf{K}} \Gamma \cup \{\neg\varphi\}$. Hence, not every model for $\Gamma$ is one for $\varphi$, i.e., $\Gamma \not\models^{\mathsf{K}} \varphi$. Strong completeness also follows by an alternative way of resaoning: note that our language is *compact*: i.e., if $\Gamma \models^{\mathsf{K}} \varphi$, there there is a finite set $\Gamma' \subseteq \Gamma$ such that $\Gamma' \models^{\mathsf{K}} \varphi$. This is seen as follows: we know from Corollary 1 that there are at most $M$ different formulas that are not provably equivalent, for an $M$ that depends on the number of agents and the number of atoms. But then, by soundness, there are also at most $M$ semantically different formulas in $\Gamma$. Putting those formulas in $\Gamma'$ gives the desired result. Strong completeness then follows from (weak) completeness with compactness.

All in all, we obtain the following (also using Lemma 2).

**Theorem 3** *The language of* DCL-PC *is compact. Moreover, the axiomatic system* DCL-PC *is sound and complete with respect to both the Kripke and the direct semantics. It is also strongly complete.*

```
function program-eval(τ, M = ⟨A, P, ξ, θ⟩, d) returns 'fail' or a model over A, P
  if τ = φ? then
    return M if DCL-PC-eval(φ, M)
       or 'fail' otherwise
  elsif τ = (i ⇝_p j) then
    return ⟨A, P, ξ′, θ⟩ if p ∈ P_i
          where ξ′ = ξ if i = j
          otherwise ξ = ⟨P_1, ..., P_n⟩ and ξ′ = ⟨P′_1, ..., P′_n⟩
          with P′_i = P_i \ {p},
             P′_j = P_j ∪ {p}, and
             P′_m = P_m for all m ≤ n, m ≠ i, j
       or 'fail' otherwise
  elsif τ = (τ_1; τ_2) then
    return program-eval(τ_2, program-eval(τ_1, M, d), d)
  elsif τ = (τ_1 ∪ τ_2) then non-deterministically choose to either
    return program-eval(τ_1, M, d)
       or program-eval(τ_2, M, d)
  elsif τ = τ′* then
    return 'fail' if d = 0
       or otherwise (if d > 0) non-deterministically choose to either
          return M
             or program-eval((τ′; τ′*), M, d − 1)
end-function
```

Figure 7: An algorithm for deciding $(\mathcal{M}, \mathcal{M}') \in R_\tau$.

*That is, for all sets of* DCL-PC *formulas* $\Gamma$ *and for every* DCL-PC *formula* $\varphi$, *we have*

$$\Gamma \vdash \varphi \ \textit{ iff } \ \Gamma \models^{\mathsf{K}} \varphi \ \textit{ iff } \ \Gamma \models^{\mathsf{d}} \varphi$$

## 4. Complexity

The model checking and satisfiability problems for CL-PC are PSPACE-complete (van der Hoek & Wooldridge, 2005b), and since DCL-PC subsumes CL-PC, this implies a PSPACE-hardness lower bound on the corresponding problems for DCL-PC. The obvious question is then whether the additional dynamic constructs of DCL-PC lead to a more complex decision problem – and in particular, whether DCL-PC satisfiability matches the EXPTIME-completeness of PDL satisfiability (Harel et al., 2000). In this section, we show that the model checking and satisfiability problems *are in fact no worse than* CL-PC: they are both PSPACE-complete. Notice that EXPTIME-completeness is usually regarded as the characteristic complexity of logics in which there is a modal operator and another operator representing the transitive closure of this operator (Blackburn et al., 2001).

Note that when we consider the model checking problem in this section, we consider the problem with respect to *direct* models, not Kripke models. Of course, with respect to satisfiability, it makes no difference: a formula is satisfiable with respect to direct models iff it is satisfiable w.r.t. Kripke models.

Before proving PSPACE-completeness for DCL-PC model checking, consider some auxiliary notions first. A *program sequence* is a transfer program that is composed of atomic transfer programs, tests, and sequential composition only. A program $\tau$ *admits* a program sequence $\beta$ if $\tau$ can be un-folded into $\beta$ by recursively applying the following rules: For any atomic transfer program $(i \rightsquigarrow_p j)$, test $\varphi?$, and transfer programs $\tau_m$ ($m \geq 0$):

$$\begin{array}{rcl} (i \rightsquigarrow_p j) & \rightsquigarrow & (i \rightsquigarrow_p j) \\ \varphi? & \rightsquigarrow & \varphi? \\ \tau_1;\tau_2 & \rightsquigarrow & \tau_1;\tau_2 \\ \tau_1 \cup \tau_2 & \rightsquigarrow & \tau_1 \text{ or } \tau_2 \\ \tau^* & \rightsquigarrow & \tau_1;\tau_2;\ldots;\tau_n, \text{ for some } n \geq 0, \\ & & \text{where } \tau_m = \tau, \text{ for all } m \leq n \end{array}$$

The following two lemmas establish that membership in the accessibility relation $R_\tau$ for a transfer program $\tau$ can be decided in polynomial space.

**Lemma 5** *For all transfer programs $\tau'$ and all (direct) models $\mathcal{M}$ and $\mathcal{M}'$, $(\mathcal{M}, \mathcal{M}') \in R_{\tau'}$ implies that $\tau'$ admits a program sequence $\beta$ of length at most exponential in the length of $\tau'$ such that $(\mathcal{M}, \mathcal{M}') \in R_\beta$. In fact, the length of $\beta$ can be limited to $2^{|\tau'|^3}$.*

**Proof:** Let $\tau'$, $\mathcal{M}$, and $\mathcal{M}'$ be as in the lemma. The proof is by induction on the structure of $\tau'$. The only interesting case is where $\tau' = \tau^*$; the other cases are straightforward. Suppose $(\mathcal{M}, \mathcal{M}') \in R_{\tau^*}$. Since $R_{\tau^*} = (R_\tau)^*$, there is a sequence $\mathcal{M}_0, \ldots, \mathcal{M}_n$, $n > 0$, of models such that $\mathcal{M}_0 = \mathcal{M}$, $\mathcal{M}_n = \mathcal{M}'$, and $(\mathcal{M}_{i-1}, \mathcal{M}_i) \in R_\tau$, for each $i$ with $1 \leq i \leq n$. By the transitivity of $R_{\tau^*}$, we can assume that the sequence $\mathcal{M}_0, \ldots, \mathcal{M}_n$ is such that $\mathcal{M}_i \neq \mathcal{M}_j$, for all $i, j$ with $1 \leq i < j \leq n$, i.e., the sequence of models contains no loops. The induction hypothesis yields that $\tau$ admits program sequences $\beta_1, \ldots, \beta_n$ such that $\beta_i$ is of length at most $2^{|\tau|^3}$ and $(\mathcal{M}_{i-1}, \mathcal{M}_i) \in R_{\beta_i}$ for each $i$ with $1 \leq i \leq n$. But then $\beta = \beta_1; \beta_2; \ldots; \beta_n$ is a program sequence admitted by $\tau^*$ such that $(\mathcal{M}, \mathcal{M}') \in R_{\tau^*}$. In the following, it is shown that $\beta$ has the required length. Note that all models reachable from $\mathcal{M} = \langle \mathbb{A}, \mathbb{P}, \xi, \theta \rangle$ via $R_{\tau^*}$ only differ in the allocation $\xi$ of propositional variables in $\mathbb{P}$ to the agents in $\mathbb{A}$. More precisely, they differ in the allocation of propositional variables to agents that occur in $\tau$. Thus there are at most $\ell^m$ such reachable models, where $\ell$ is the number of propositional variables occurring in $\tau$ and $m$ the number of agents occurring in $\tau$. Notice that $n$ does not exceed $\ell^m$; otherwise the sequence $\mathcal{M}_0, \ldots, \mathcal{M}_n$ contains loops contradicting the assumption. Together with the fact that $\ell^m \leq |\tau|^{|\tau|} \leq 2^{|\tau|^2}$, an upper bound for the length of $\beta$ can be given as follows:

$$\begin{array}{rcl} |\beta| = |\beta_1; \beta_2; \ldots; \beta_n| & \leq & n \times \sup\{|\beta_i| : 1 \leq i \leq n\} + n \\ & \leq & 2^{|\tau|^2} \times 2^{|\tau|^3} + 2^{|\tau|^2} \\ & = & 2^{|\tau|^2 + |\tau|^3} + 2^{|\tau|^2} \\ & \leq & 2^{(|\tau|+1)^3} \\ & \leq & 2^{|\tau^*|^3}. \end{array}$$

QED

**Lemma 6** *For all programs $\tau$ and all (direct) models $\mathcal{M}$ and $\mathcal{M}'$, the membership problem $(\mathcal{M}, \mathcal{M}') \in R_\tau$ can be decided in* PSPACE.

**Proof:** Let $\tau$ be a program and let $\mathcal{M}$, $\mathcal{M}'$ be two (direct) models. Consider the following algorithm that decides $(\mathcal{M}, \mathcal{M}') \in R_\tau$ by using the function *program-eval*$(\cdots)$ in Figure 7:

1. Set $d = 2^{|\tau|^3}$.

2. If *program-eval*$(\tau, \mathcal{M}, d) = \mathcal{M}'$, then return '$(\mathcal{M}, \mathcal{M}') \in R_\tau$', and 'No' otherwise.

To see that this algorithm is correct, it is shown that *program-eval*$(\tau, \mathcal{M}, d) = \mathcal{M}'$ iff $(\mathcal{M}, \mathcal{M}') \in R_\tau$. For the direction from left to right, it is readily checked that *program-eval*$(\tau, \mathcal{M}, d) = \mathcal{M}'$ implies the existence of a program sequence $\beta$ admitted by $\tau$ of length at most $|\tau| \times d$ such that $(\mathcal{M}, \mathcal{M}') \in R_{\tau'}$. Clearly, $R_\beta \subseteq R_\tau$ and thus $(\mathcal{M}, \mathcal{M}') \in R_\tau$. Consider the direction from right to left. From $(\mathcal{M}, \mathcal{M}') \in R_\tau$, it follows by Lemma 5 that there is a program sequence $\beta$ admitted by $\tau$ of length at most $2^{|\tau|^3}$ such that $(\mathcal{M}, \mathcal{M}') \in R_\beta$. Step 1 ensures that the value of $d$ is such that $2^{|\tau|^3} \leq |\tau| \times d$. Then it is obvious by construction of the algorithm that the non-deterministic choices in the lines 15 and 20 of Figure 7 yield that *program-eval*$(\tau, \mathcal{M}, d) = \mathcal{M}'$. Notice that the algorithm terminates since the recursive calls in the lines 14, 16, and 17 are applied on strict subprograms only and the recursive call in Line 22 is followed by the one in Line 14 while the parameter $d$ limits the recursion depth.

The above algorithm can be run in polynomial space. To see this, notice that the function DCL-PC-*eval*$(\cdots)$, which is called in Line 3, can be computed in polynomial space and that the parameter $d$ is encoded in binary. Moreover, the stack of an algorithm computing the function *program-eval*$(\cdots)$ can be limited to a size polynomial in the length of $\tau$. Note that the stack only needs to store the currently evaluated program and the programs at the backtracking points, which are introduced at the nested function call in Line 14. But since this nested function call is applied on strict subprograms, there are only linearly many backtracking points needed at a time. Although the algorithm is non-deterministic, it follows from the well-known fact NPSPACE equals PSPACE (Savitch, 1970) that it runs in PSPACE. QED

Using the previous two lemmas, we can now prove the following.

**Theorem 4** *The model checking problem* DCL-PC *(w.r.t. direct models) is* PSPACE*-complete.*

**Proof:** Given that DCL-PC subsumes the PSPACE-hard logic CL-PC, we only need to prove the upper bound. Consider the function DCL-PC-*eval*$(\cdots)$ in Figure 8. Soundness is obvious by construction. First note that the algorithm is strictly analytic: recursion is always on a sub-formula of the input. That the algorithm is in PSPACE follows from the fact that the loops at lines 10–12 and 15-18 involve, in the first case simply binary counting with the variables $\mathbb{P}_C$, and in the second simply looping through all direct models over $\mathbb{A}$ and $\mathbb{P}$: we do not need to store these models once they are checked, and so this can be done in polynomial space. Finally, Lemma 6 yields that the check $(\mathcal{M}, \mathcal{M}') \in R_\tau$ on Line 16 can be done in polynomial space. QED

Now, we make use of the following result, the proof of which is identical to the equivalent result proved by van der Hoek and Wooldridge (2005b).

**Lemma 7** *If a* DCL-PC *formula $\varphi$ is satisfiable, then it is satisfied in a (direct) model $\mathcal{M}$ such that size$(\mathcal{M}) = |\mathbb{P}(\varphi)| + |Ag(\varphi)| + 1$.*

We can now prove the following.

**Theorem 5** *The satisfiability checking problem for* DCL-PC *is* PSPACE*-complete.*

```
function DCL-PC-eval($\varphi, \mathcal{M} = \langle \mathbb{A}, \mathbb{P}, \xi_0, \theta \rangle$) returns tt or ff
    if $\varphi \in \mathbb{P}$ then
        return $\theta(\varphi)$
    elsif $\varphi = \neg\psi$ then
        return not DCL-PC-eval($\psi, \mathcal{M}$)
    elsif $\varphi = \psi_1 \vee \psi_2$ then
        return DCL-PC-eval($\psi_1, \langle \mathbb{A}, \mathbb{P}, \xi_0, \theta \rangle$)
            or DCL-PC-eval($\psi_2, \langle \mathbb{A}, \mathbb{P}, \xi_0, \theta \rangle$)
    elsif $\varphi = \Diamond_C \psi$ then
        for each C-valuation $\theta_C$
                if DCL-PC-eval($\psi, \langle \mathbb{A}, \mathbb{P}, \xi_0, \theta \rangle \oplus \theta_C$) then return tt
        end-for
        return ff
    elsif $\varphi = \langle \tau \rangle \psi$ then
        for each model $\mathcal{M}'$ over $\mathbb{A}$, $\mathbb{P}$
            if $(\mathcal{M}, \mathcal{M}') \in R_\tau$ then
                if DCL-PC-eval($\varphi, \mathcal{M}'$) then return tt
        end-for
        return ff
end-function
```

Figure 8: A model checking algorithm for DCL-PC.

**Proof:** Given a formula $\varphi$, loop through each model $\mathcal{M}$ containing $\mathbb{P}(\varphi)$ and $Ag(\varphi)$ such that $size(\mathcal{M}) = |\mathbb{P}(\varphi)| + |Ag(\varphi)| + 1$, and if $\mathcal{M} \models^{\mathsf{d}} \varphi$ then return 'Yes'. If we have considered all such models, return 'No'. By Theorem 4, we can check whether $\mathcal{M} \models^{\mathsf{d}} \varphi$ in polynomial space. QED

Notice that the PSPACE complexity for checking satisfiability depends upon the fact that models for DCL-PC are concise, and that hence we can loop through them all in polynomial space (we do not need to 'remember' a model after it has been considered).

## 5. Characterizing Control

One of the main concerns in the original study of CL-PC (van der Hoek & Wooldridge, 2005b) was to investigate the logical characterization of *control*: the extent to which we could characterize, in the logic, what states of affairs agents could reliably control. Control was distinguished from ability in the sense that, for example, no agent could be said to control a tautology, even if one might be prepared to concede that an agent would have the ability to bring about a tautology. The starting point for the study of control (van der Hoek & Wooldridge, 2005b) was the $controls(i, p)$ construct: as we have already seen, such an expression will be true iff the variable $p$ is under the control of agent $i$. This led to an analysis and characterization of the types of formulas that an agent could be said to control. The type of control studied by van der Hoek and Wooldridge derives from the ability of agents to choose values for the propositional variables under their control. Let us refer to this type of control, where an agent is directly able to exert some influence over some state of affairs by assigning values to its variables, as *first-order control*. In this section, we undertake a similar study of control in the richer setting of DCL-PC. Here, however, we have a second type of control, which derives from the ability to *transfer control of variables to other agents*. Thus, for example, if $i$ controls $p$, she also 'has the power' to ensure for instance $controls(j,p)$, where $j$ is an agent different from $i$. This control is expressed through the transfer modality: $\langle i \rightsquigarrow_p j\rangle controls(j,p)$. We refer to this type of control as *second-order control*. We will see that these types of 'control' are indeed rather orthogonal. For instance, $\langle i \rightsquigarrow_p j\rangle \Diamond_j \varphi$ ($i$ can give $p$ to $j$, who then can achieve $\varphi$) and $\Diamond_{i,j}\varphi$ ($i$ and $j$ can cooperate, to achieve $\varphi$) are logically incomparable. For example, taking $\varphi = \langle j \rightsquigarrow_p i\rangle\top$ gives

$$\models^{\mathsf{d}} controls(i,p) \rightarrow (\langle i \rightsquigarrow_p j\rangle\varphi \wedge \neg\Diamond_{i,j}\varphi)$$

while for $\varphi = \langle i \rightsquigarrow_p j\rangle\top$ and assuming $i \neq j$, we have

$$\models^{\mathsf{d}} controls(i,p) \rightarrow (\neg\langle i \rightsquigarrow_p j\rangle\varphi \wedge \Diamond_{i,j}\varphi).$$

However, if the goal is an objective formula, we can relate atomic control and transfer, as we will shortly see.

To begin our study, consider the transfer program

$$\texttt{give}_i \mathrel{\hat{=}} \bigcup_{p\in\mathbb{P}} \big(controls(i,p)?; \bigcup_{j\in\mathbb{A}} i \rightsquigarrow_p j\big). \tag{20}$$

Then $\langle \texttt{give}_i\rangle\varphi$ would express that $i$ has a way to give one of her propositional variables to one of the agents (possibly herself) in such a way that consequently $\varphi$ holds. Thus, $\langle \texttt{give}_i^*\rangle\varphi$ means that $i$ can distribute her variables among the agents in such a way that afterwards $\varphi$ holds. Hence, when reasoning about $i$'s power, the strongest that she can achieve is any $\varphi$ for which $\Diamond_i\varphi \vee \langle \texttt{give}_i^*\rangle\varphi$, expressing that $i$ can achieve $\varphi$ by either choosing an appropriate value for her variables, or by distributing her variables over $\mathbb{A}$ in an appropriate way. Note that both $\Diamond_i\varphi$ and $\langle \texttt{give}_i^*\rangle\varphi$ imply $\langle \texttt{give}_i^*\rangle\Diamond_i\varphi$, and hence any $\varphi$ for which $\langle \texttt{give}_i^*\rangle\Diamond_i\varphi$ holds can be seen as what $i$ can achieve on her own. We will come back to the program $\texttt{give}_i$ below.

The program `give` can be generalized to incorporate coalitions that can give away variables, and those that can receive: let

$$\texttt{give}_{C\rightsquigarrow D} \mathrel{\hat{=}} \bigcup_{i\in C}\bigcup_{p\in\mathbb{P}} \big(controls(i,p)?; \bigcup_{j\in D\cup\{i\}} \langle i \rightsquigarrow_p j\rangle\big). \tag{21}$$

This program $\texttt{give}_{C\rightsquigarrow D}$ lets an arbitrary agent $i$ from the coalition $C$ either give any of her variables $p$ to an arbitrary member of the coalition $D$, or do nothing (i.e., give them to herself). Now, for objective formulas $\varphi$, we have the following, where $i$ is a dedicated agent from $C$:

$$\Diamond_C\varphi \leftrightarrow \langle \texttt{give}^*_{C\rightsquigarrow\{i\}}\rangle\Diamond_i\varphi.$$

In words: the agents in the coalition $C$ can choose values for their variables such that $\varphi$, if and only if they have a way to give all their variables to the dedicated agent $i$, who then can achieve $\varphi$. Note that we are in general not able to eliminate all occurrences of $\Diamond$'s, since this is the only way to express first-order control, i.e., to reason about a 'different valuation'.

For some examples in the language without transfer, we refer to a paper by van der Hoek and Wooldridge (2005b), especially to the example 'Bach or Stravinsky', (i.e., Example 2.4, van der Hoek & Wooldridge, 2005b). Before looking at two examples of control in a dynamic setting, note that $\Diamond_{\mathbb{A}}$ allows the following inference, for any objective formula $\varphi$:

$$\varphi \text{ is consistent } \Rightarrow \models^{\mathsf{d}} \Diamond_{\mathbb{A}}\varphi \tag{22}$$

This inference says that the grand coalition $\mathbb{A}$ can achieve any satisfiable objective formula.

**Example 1** *Suppose we have n agents: 1, ..., n. Each controls a flag $r_i$ ($i = 1 \ldots n$) to indicate that they desire control over a particular resource, modeled with a variable p. It is not that they want p to be true or false every now and then, (which could be taken care of by a central agent executing a program making p false and true alternatively), but rather, they want to* control *p eventually. Let $+_n$ denote addition modulo n, and, similarly, $-_n$ subtraction modulo n. Let* skip *denote $\top?$, i.e., a test on a tautology. Consider the following program:*

$$\begin{array}{llllll} \texttt{grant-req}(i) \hat{=} & \texttt{if} & \neg controls(i,p) & \texttt{then} & \texttt{skip else} \\ & \texttt{if} & r_{i+_n 1} & \texttt{then} & (i \rightsquigarrow_p i +_n 1)\ \texttt{else} \\ & \cdots & & \cdots & \\ & \texttt{if} & r_{i+_n(n-1)} & \texttt{then} & (i \rightsquigarrow_p i +_n (n-1))\ \texttt{else skip} \end{array}$$

*The program* $\texttt{grant-req}(i)$ *makes agent i pass on the resource p whenever she has it and somebody else needs it, where the need is checked in the order starting with the agent with the next $+_n$ index. Note that the 'use' of this variable p, i.e., making it true or false, is not encoded in our program constructs. Now consider the program*

$$\texttt{pass-on}(i,j) \hat{=} \texttt{grant-req}(i); \ldots; \texttt{grant-req}(j -_n 1).$$

*The program* $\texttt{pass-on}(i,j)$ *will pass control over the variable p to agent j, provided that initially $r_j$ is set and one of the agents in the sequence $i, i +_n 1, \ldots, j -_n 1$ owns it. This can be expressed as follows:*

$$r_j \wedge controls(\{i, i +_n 1, \ldots, j -_n 1\}, p) \rightarrow [\texttt{pass-on}(i,j)]controls(j,p).$$

*Now we have:*

$$r_i \rightarrow \big(\langle \texttt{pass-on}(i +_n 1, i)\rangle controls(i,p) \\ \wedge [\texttt{pass-on}(i +_n 1, i)]controls(i,p)\big).$$

*That is: if agent i flags a request $r_i$ for resource p, then, after the program* $\texttt{pass-on}(i +_n 1, i)$ *has been executed, i will be under control of p.*

Notice that the previous example 'freely' passes on a variable along chains of agents, thereby taking for granted that they can control that variable on the fly, and making it true or false at will. In the following example, control over a variable is not only important, but also the truth of some side conditions involving them.

**Example 2** *We have a scenario with three agents: two clients $c_1$ and $c_2$, and a server s. The server always has control over one of the propositional variables $p_1$ and $p_2$, in particular s wants to guarantee that those variables are never true simultaneously. At the same time, $c_1$ and $c_2$ want to ensure that at least one of the variables $p_i$ ($i = 1, 2$) is true, where variable $p_i$ belongs to client $c_i$. We can describe the invariant of the system with the formula Inv:*

$$Inv \hat{=} \bigvee_{i=1,2} controls(s, p_i) \ \wedge \ \bigvee_{i=1,2} controls(c_i, p_i)$$

*Consider the following transfer program $\beta$:*

$$\beta \hat{=} \ \big((controls(s,p_1)? \ ; \ s \rightsquigarrow_{p_1} c_1 \ ; \ c_2 \rightsquigarrow_{p_2} s) \\ \cup (controls(s,p_2)? \ ; \ s \rightsquigarrow_{p_2} c_2 \ ; \ c_1 \rightsquigarrow_{p_1} s)\big)^*.$$

*This says that an arbitrary number of times one variable $p_i$ is passed from the server to the client $c_i$, and another variable $p_j$ ($i \neq j$) from the client $c_j$ to the server.*

*Using Inv and $\beta$, we can describe the whole scenario as follows:*

$$Inv \rightarrow [\beta]Inv$$
$$Inv \rightarrow [\beta]\big(\Diamond_s\neg(p_1 \wedge p_2) \wedge \Diamond_{\{c_1,c_2\}}(p_1 \vee p_2)\big)$$

A general characterization of the types of formulas that agents and coalitions could control was given by van der Hoek and Wooldridge (2005a), and our aim is now to undertake the same study for DCL-PC. It will appear that this can be done on a local and a global level, but we will also see that the notion of control that we inherited from CL-PC, has a natural generalization in our context.

The next corollary establishes a result concerning the characterization of control. Its first item says that of no strict sub-coalition $C \neq \mathbb{A}$ of $\mathbb{A}$, is it valid that $C$ controls something. In other words, control of such a coalition is always a feature of a specific model, in particular, a specific allocation. According to the second item, the grand coalition $\mathbb{A}$ derivably, or in all models, controls exactly those formulas $\varphi$ with the property that their equivalent form $\bigvee_{\upsilon\in\Upsilon}(\bigvee \Pi_\upsilon(\varphi) \wedge \upsilon)$ is such that for every allocation description $\upsilon$, the formula $\bigvee \Pi_\upsilon(\varphi)$ is a contingency, i.e., not a tautology and neither a contradiction. If $\varphi$ is a propositional formula, we have $\varphi \leftrightarrow \bigvee \Pi_\upsilon(\varphi)$ and it is easy to see that $\varphi$ being a contingency is sufficient and necessary to have $\models^{\mathsf{K}}$ *controls*$(\mathbb{A},\varphi)$. On the other hand, if $\varphi$ is for instance *controls*$(i,p)$, then $\Pi_\upsilon(\varphi) = \Pi$ if $\upsilon \rightarrow$ *controls*$(i,p)$ and $\bigvee \Pi_\upsilon(\varphi)$ equals $\top$. And indeed, $\not\models^{\mathsf{K}}$ *controls*$(\mathbb{A},\varphi)$. Contrast this with $\varphi = p \wedge$ *controls*$(i,p)$. If follows easily from the truth definition of the Kripke semantics defined in Section 2.6, and from Theorem 2 and Theorem 3.

**Corollary 2** *Let $\mathfrak{M}$ be a Kripke model, C any coalition $C \subseteq \mathbb{A}$ with $C \neq \mathbb{A}$, and let $\varphi$ be ranging over* DCL-PC *formulas. Then it follows that:*

1. *For no $\varphi$, do we have $\models^{\mathsf{K}}$ controls$(C,\varphi)$, and*

2. *$\models^{\mathsf{K}}$ controls$(\mathbb{A},\varphi)$ iff the formula $\bigvee_{\upsilon\in\Upsilon}(\bigvee \Pi_\upsilon(\varphi) \wedge \upsilon)$, to which $\varphi$ is equivalent according to Theorem 2, is such that for no allocation description $\upsilon$, $\Pi_\upsilon(\varphi) = \Pi$ or $\Pi_\upsilon(\varphi) = \emptyset$.*

**Proof:**

1. In order for *controls*$(C,\varphi)$ to be valid under the Kripke semantics, it has to be true at all worlds in all Kripke models. Take any Kripke model $\mathfrak{M} = \langle\Theta, R_{i\in\mathbb{A}}, \xi\rangle$ for which the allocation $\xi = \langle\mathbb{P}_1,\mathbb{P}_2,\ldots,\mathbb{P}_n\rangle$ is such that $\mathbb{P}_j = \emptyset$, for all $j \in C$. We then have $\mathfrak{M},\theta \models^{\mathsf{K}} \Diamond_C\varphi$ iff for some world $\theta' \in \Theta$ with $\theta' = \theta \pmod{\mathbb{P}_C}$, it holds that $\mathfrak{M},\theta' \models^{\mathsf{K}} \varphi$. But, since $\mathbb{P}_C = \emptyset$, the only such $\theta'$ is $\theta$ itself, so that we cannot have $\mathfrak{M},\theta \models^{\mathsf{K}} \Diamond_C\varphi \wedge \Diamond_C\neg\varphi$. Hence, $\mathfrak{M},\theta \not\models^{\mathsf{K}}$ *controls*$(C,\varphi)$.

2. First we prove the left-to-right direction by contraposition. Let $\varphi$ be equivalent to the formula $\bigvee_{\upsilon\in\Upsilon}(\bigvee \Pi_\upsilon(\varphi) \wedge \upsilon)$, which we have by Theorem 2. Suppose, for some allocation description $\upsilon$, that $\Pi_\upsilon(\varphi) = \Pi$. This means, for every Kripke model $\mathfrak{M} = \langle\Theta, R_{i\in\mathbb{A}}, \xi_\upsilon\rangle$ and for every valuation $\theta$, that $\mathfrak{M},\theta \models^{\mathsf{K}} \varphi$. Consequently, $\mathfrak{M},\theta \models^{\mathsf{K}} \Diamond_{\mathbb{A}}\varphi$. However, the agents in $\mathbb{A}$ can only change $\theta$, but not the current allocation $\xi_\upsilon$. Since $\Pi_\upsilon(\varphi) = \Pi$, $\mathbb{A}$ cannot choose a valuation that falsifies $\varphi$, i.e., $\mathfrak{M},\theta \models^{\mathsf{K}} \neg\Diamond_{\mathbb{A}}\neg\varphi$. Similarly, if $\Pi_\upsilon(\varphi) = \emptyset$, we have $\mathfrak{M},\theta \models^{\mathsf{K}} \Diamond_C \neg\varphi$, for each $\theta$. But, given $\upsilon$, $\mathbb{A}$ cannot choose a valuation satisfying $\varphi$ on the current allocation described by $\upsilon$, i.e., $\mathfrak{M},\theta \models^{\mathsf{K}} \neg\Diamond_{\mathbb{A}}\varphi$. Hence, in either case we have $\not\models^{\mathsf{K}}$ *controls*$(\mathbb{A},\varphi)$.

Consider the other direction from right to left. Suppose $\varphi$ is equivalent to $\bigvee_{\upsilon\in\Upsilon}(\bigvee \Pi_\upsilon(\varphi)\wedge\upsilon)$ while, for no $\upsilon$, the set $\Pi_\upsilon(\varphi)$ is either $\Pi$ or $\emptyset$. Let $\mathfrak{M} = \langle\Theta, R_{i\in\mathbb{A}}, \xi\rangle$ be a Kripke model. Remember that $\upsilon_\xi$ is the allocation description corresponding to the allocation $\xi$. By the fact that $\Pi_{\upsilon_\xi} \neq \emptyset$, there is a valuation description $\pi \in \Pi_{\upsilon_\xi}$ such that the corresponding valuation $\theta_\pi$ satisfies $\varphi$ at $(\mathfrak{M},\theta_\pi)$. But then, $\mathbb{A}$ can choose $\theta_\pi$ in order to satisfy $\varphi$, and thus we have $\mathfrak{M},\theta \models^{\mathsf{K}} \Diamond_{\mathbb{A}}\varphi$, for any $\theta$. Similarly, we have $\mathfrak{M},\theta \models^{\mathsf{K}} \Diamond_{\mathbb{A}}\neg\varphi$ which follows by the fact that $\Pi_{\upsilon_\xi} \neq \Pi$. Hence, we have $\models^{\mathsf{K}}$ *controls*$(\mathbb{A},\varphi)$.

QED

One may ask for a more local characterization of what a coalition controls: for which Kripke models $\mathfrak{M}$ and valuations $\theta$ do we have $\mathfrak{M},\theta \models^{\mathsf{K}}$ *controls*$(C,\varphi)$? For this notion of control, the answer can be immediately read off from Theorem 6, to be given shortly. That theorem is about a more general notion: to recover a characterization result for the current notion *controls*$(i,\varphi)$, we would only need the items (1b) and (2b) of Theorem 6.

The notion of control discussed so far is that taken from CL-PC: we have lifted the characterization results to our richer language. However, as is clear from our discussion earlier in this section, a more appropriate notion of control of an individual $i$ in our language might be obtained using the program $\texttt{give}_i^*$, where $\texttt{give}_i$ is defined in (20). Note that $\Diamond_i\varphi \rightarrow \langle\texttt{give}_i^*\rangle\Diamond_i\varphi$ is valid, and hence that $\langle\texttt{give}_i^*\rangle$ seems a more general way to reason about $i$'s control: it is about what $i$ can achieve *both by toggling its propositional variables and delegating some of them*. One can easily discuss this at the coalitional level, by lifting Definition (20) to the case of coalitions $C$ as it was suggested in (21) with $\texttt{give}_{C\leadsto D}$. However, we stick to the individual case here for simplicity. Let us therefore define

$$CONTROLS(i,\varphi) \hat{=} \left(\langle\texttt{give}_i^*\rangle\Diamond_i\varphi \wedge \langle\texttt{give}_i^*\rangle\Diamond_i\neg\varphi\right) \quad (23)$$

This definition says that agent $i$ controls a formula $\varphi$ iff there is a way to distribute her propositional variables over the agents such that after $i$ makes appropriate choices for her remaining variables, $\varphi$ holds, but there is also a way of distributing her variables that enables her to enforce $\neg\varphi$. From the validity of $\Diamond_i\varphi \rightarrow \langle\texttt{give}_i^*\rangle\Diamond_i\varphi$, we infer that *controls*$(i,\varphi)$ implies *CONTROLS*$(i,\varphi)$. Notice that the implication the other way around is not valid since *controls*$(i,\cdot)$ can never be true for control of other agents over variables. For example, *CONTROLS*$(i,$ *controls*$(j,p))$ holds iff $p \in \mathbb{P}_i$. From this, we know that *controls*$(i,p) \rightarrow$ *CONTROLS*$(i,$ *controls*$(j,p))$ is a theorem, which basically says that when having control over a variable, you can freely choose to keep it or pass it on. However, *controls*$(i,p) \rightarrow$ *controls*$(i,$ *controls*$(j,p))$ is not valid, we even have *controls*$(i,p) \rightarrow \neg$*controls*$(i,$ *controls*$(j,p))$: once agent $i$ owns $p$, she cannot choose to keep $p$ or to pass it on by only toggling her propositional variables.

Before we state our characterization result, we introduce some more notation. For any two Kripke models $\mathfrak{M} = \langle\Theta, R_{i\in\mathbb{A}}, \xi\rangle$ and $\mathfrak{M}' = \langle\Theta, R_{i\in\mathbb{A}}, \xi'\rangle$ in $\mathcal{K}(\mathbb{A},\mathbb{P})$ and for any agent $i$, we say that $\mathfrak{M} \geq_i \mathfrak{M}'$ if the allocations $\xi = \langle\mathbb{P}_1,\ldots,\mathbb{P}_i,\ldots,\mathbb{P}_n\rangle$ and $\xi' = \langle\mathbb{P}'_1,\ldots,\mathbb{P}'_i,\ldots,\mathbb{P}'_n\rangle$ are such that $\mathbb{P}'_i \subseteq \mathbb{P}_i$ and, for all $j \neq i$, $\mathbb{P}_j \subseteq \mathbb{P}'_j$. That is, $\mathfrak{M}'$ is obtained from $\mathfrak{M}$ by executing $\texttt{give}_i^*$. In such a case, we also say $\xi \geq_i \xi'$.

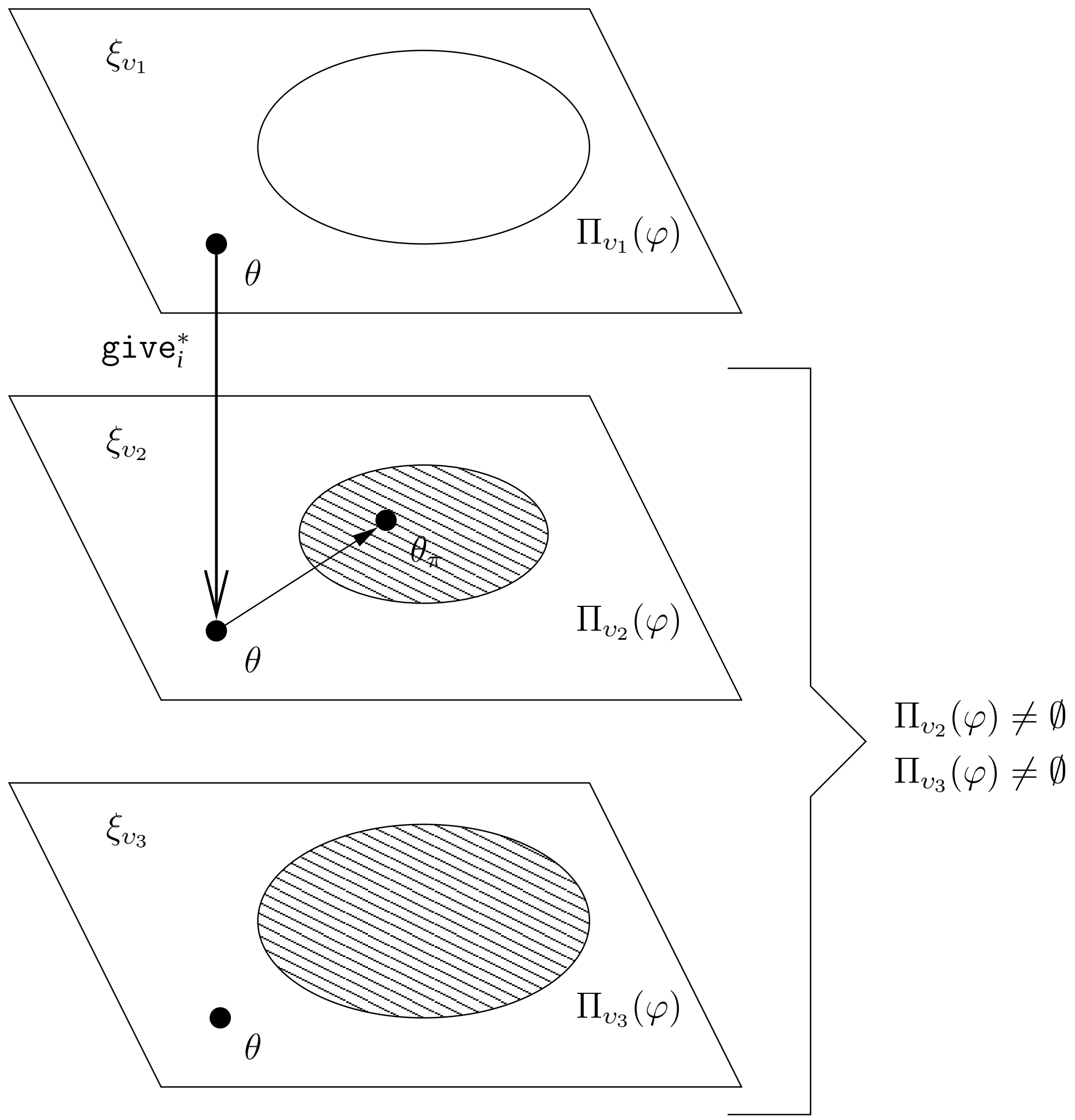

Figure 9: Illustration of $\varphi \equiv \bigvee_{\upsilon\in\Upsilon} \left(\bigvee \Pi_\upsilon(\varphi) \wedge \upsilon\right)$.

For each valuation description $\pi \in \Pi$, let $\theta_\pi$ be the valuation that is described by $\pi$, and, for each allocation description $\upsilon \in \Upsilon$, let $\xi_\upsilon$ be the allocation described by $\upsilon$, and let $\mathbb{P}_i^\upsilon$ be the set of propositional variables controlled by agent *i* in $\xi_\upsilon$.

**Theorem 6** *Let $\varphi$ be a* DCL-PC *formula with $\varphi \equiv \bigvee_{\upsilon\in\Upsilon} \left(\bigvee \Pi_\upsilon(\varphi) \wedge \upsilon\right)$, as given by Theorem 2. Let $\mathfrak{M} = \langle \Theta, R_{i\in\mathbb{A}}, \xi\rangle$ be a Kripke model of $\mathcal{K}(\mathbb{A},\mathbb{P})$, and $\theta$ a world in $\mathfrak{M}$. Then, for each agent $i \in \mathbb{A}$,*

$$\mathfrak{M}, \theta \models^{\mathsf{K}} \mathit{CONTROLS}(i, \varphi)$$

*iff the following two conditions are satisfied:*

1. *There is a $\upsilon \in \Upsilon$ and a $\pi \in \Pi_\upsilon(\varphi)$ such that*

   *(a) $\xi \geq_i \xi_\upsilon$, and*

   *(b) $\theta_\pi = \theta \pmod{\mathbb{P}_i^\upsilon}$.*

2. *There is a* $\upsilon \in \Upsilon$ *and a* $\pi \in \Pi \setminus \Pi_\upsilon(\varphi)$ *such that*

   (a) $\xi \geq_i \xi_\upsilon$, *and*

   (b) $\theta_\pi = \theta \pmod{\mathbb{P}_i^\upsilon}$.

We first demonstrate the requirements (1) and (2) of Theorem (6). Suppose that $\theta$ satisfies $(p \wedge \neg q \wedge r)$, Agent $i = 1$ owns $p$ in $\mathfrak{M}$, Agent 2 owns $q$ and $r$, and Agent 3 has no propositional variables in $\mathfrak{M}$. First of all, to see why Item (1b) is needed, we have to guarantee that for some $\mathfrak{M}', \theta'$, it holds that $\mathfrak{M}', \theta' \models^{\mathsf{K}} \Diamond_1 \varphi$. That means that, even after 1 has given away some of her atoms (resulting in some allocation $\xi'$), she still should be able to make $\varphi$ true. This is possible for $\varphi = (\neg p \wedge \neg q \wedge r)$: Agent 1 could simply stay within the current allocation $\xi$ and just make $p$ false. However, this is not possible for $\varphi = (\neg p \wedge \neg q \wedge r \wedge \mathit{controls}(3, p))$ since, once 1 has delegated control of $p$ to 3, agent 1 cannot make $p$ false anymore. Moreover, an agent $i$ can only give atoms away, so any model with allocation $\xi_\upsilon$ that makes it possible for her to satisfy $\varphi$ should be one for which $\xi \geq_i \xi_\upsilon$, which explains Item (1a). Item (2a) has exactly the same motivation, and requirement (2b) is easily understood to be similar to (1b), once one realizes that the normal form of $\neg\varphi$ can be expressed in terms of the normal form of $\varphi$ as follows:

$$\neg\varphi \equiv \bigvee_{\upsilon \in \Upsilon} \big( \bigvee (\Pi \setminus \Pi_\upsilon(\varphi)) \wedge \upsilon \big).$$

For a simple illustrating example, suppose there are only two allocations $\xi_1$ and $\xi_2$, and $\varphi$ is equivalent to

$$\begin{array}{ll} & ((p \wedge q) \vee (p \wedge \neg q)) \wedge \xi_1 \\ \vee & ((\neg p \wedge q) \vee (p \wedge \neg q)) \wedge \xi_2. \end{array}$$

Note that $\varphi$'s normal form describes the valuations on $\xi_1$ and $\xi_2$ where $\varphi$ is satisfied. The normal form of $\neg\varphi$ is complementary to the one of $\varphi$ in the sense that it describes the valuations on $\xi_1$ and $\xi_2$ where $\varphi$ is falsified:

$$\begin{array}{ll} & ((\neg p \wedge \neg q) \vee (\neg p \wedge q)) \wedge \xi_1 \\ \vee & ((\neg p \wedge \neg q) \vee (p \wedge q)) \wedge \xi_2. \end{array}$$

**Proof:** We illustrate our proof with a pictorial story that shows why the requirements 1 and 2 of the theorem are both sufficient and necessary. Given that $\varphi$ is equivalent to $\bigvee_{\upsilon \in \Upsilon}(\bigvee \Pi_\upsilon(\varphi) \wedge \upsilon)$ by Theorem 2, semantically this means that $\varphi$ corresponds to a collection of the 'shaded areas', as depicted in Figure 9. Now, for $\mathit{CONTROLS}(i, \varphi)$ to be true at a world $\theta$ in a Kripke model $\mathfrak{M} = \langle \Theta, R_{i \in \mathbb{A}}, \xi \rangle$, Agent $i$ has to be able to move *inside* such a shaded area, and to move *outside* it as well. But moving inside a shaded area, means being able to first go to a model with allocation $\xi_\upsilon$, and then to a world $\theta_\pi$ within that model such that the valuation description $\pi$ is in $\Pi_\upsilon$. Notice that $i$ can move to allocation $\xi_\upsilon$ only by delegating control over her variables to other agents (hence the requirement $\xi \geq_i \xi_\upsilon$), and $i$ can move to valuation $\theta_\pi$ only by toggling her remaining variables in $\mathbb{P}_i^\upsilon$ at $\xi_\upsilon$ (hence the condition $\theta_\pi = \theta \pmod{\mathbb{P}_i^\upsilon}$). This shows that Condition 1 is equivalent to $\mathfrak{M}, \theta \models^{\mathsf{K}} \langle \mathtt{give}_i^* \rangle \Diamond_i \varphi$. Accordingly, Condition 2 corresponds to being able to move *outside* of a shaded area in Figure 9. Semantically, this means being able to first go to a model with allocation $\xi_{\upsilon'}$, and then to a world $\theta_{\pi'}$ within that model such that the valuation description $\pi'$ is not in $\Pi_{\upsilon'}$. Consequently, Condition 2 is equivalent to $\mathfrak{M}, \theta \models^{\mathsf{K}} \langle \mathtt{give}_i^* \rangle \Diamond_i \neg\varphi$, which finishes the proof. QED

## 6. Possible Extensions and Refinements

In this section, we consider some possible extensions and refinements to the framework we have presented in this paper. We do not claim to have substantial results relating to these extensions – the aim is simply to indicate some possible directions for future research.

### 6.1 Separating First- and Second-Order Control

In DCL-PC as presented here, an agent can assign a value to a variable (exercising first-order control) iff it can 'give this variable away' (exercising second-order control). That is, for any pair of agents $i, j$ and propositional variable $p$, we have the following.

$$\models^{\mathsf{d}} controls(i,p) \leftrightarrow \langle i \rightsquigarrow_p j \rangle \top \tag{24}$$

A moment's reflection should confirm that this is not always how things work in human societies. We might empower an individual to make a choice on our behalf, but we might not be happy with the idea that this individual could in turn transfer this power to somebody else. Sometimes, it might be acceptable; but certainly not in all cases.

We can straightforwardly distinguish between these situations by extending our models and modifying the semantics of our language as follows. A model $\mathcal{M}$ is now defined to be a structure:

$$\mathcal{M} = \langle \mathbb{A}, \mathbb{P}, \rho, \xi_0, \theta \rangle$$

where the components $\mathbb{A}, \mathbb{P}, \xi_0$, and $\theta$ are as originally defined, and $\rho = \langle \rho_1, \ldots, \rho_n \rangle$ is a tuple of subsets of $\mathbb{P}$, with elements indexed by agents $\mathbb{A}$, such that $\rho_1, \ldots, \rho_n$ forms a partition of $\mathbb{P}$.

Now, the intended interpretation of such models is as follows:

- the partition $\xi_0$ defines who (initially) has the ability to assign values to which variables (i.e., who has first-order control of variables); while
- the partition $\rho$ defines who can transfer control of which variables (i.e., who has second-order control of variables).

Syntactically, the logic that we define to reason about such structures is identical to DCL-PC, however the semantics are different. In fact, the only element of the semantics that we need to change relates to the definition of the accessibility relation for atomic transfer programs.

Let $\mathcal{M} = \langle \mathbb{A}, \mathbb{P}, \rho, \xi_0, \theta \rangle$ and $\mathcal{M}' = \langle \mathbb{A}, \mathbb{P}, \rho, \xi'_0, \theta \rangle$ be two models with $\xi_0 = \langle \mathbb{P}_1, \ldots, \mathbb{P}_n \rangle$ and $\xi'_0 = \langle \mathbb{P}'_1, \ldots, \mathbb{P}'_n \rangle$. Then:

$$(\mathcal{M}, \mathcal{M}') \in R_{i \rightsquigarrow_p j}$$

iff

1. $p \in \rho_i$ (agent $i$ has second-order control of $p$ to begin with)
2. for all $k \in \mathbb{A}$, if $p \in \mathbb{P}_k$ then:
   (a) if $k = j$ then $\forall l \in \mathbb{A}$, $\mathbb{P}_l = \mathbb{P}'_l$.
   (b) if $k \neq j$ then:

- $\mathbb{P}'_j = \mathbb{P}_j \cup \{p\}$,
- $\mathbb{P}'_k = \mathbb{P}_k \setminus \{p\}$,
- $\forall l \in \mathbb{A} \setminus \{j, k\}, \mathbb{P}_l = \mathbb{P}'_l$.

With this setup, first-order control is dynamic, and can be changed by transfer programs, while second-order control as defined in $\rho$ is static. Moreover, the fact that an agent has first-order control of a variable does not mean it has second-order control: we no longer have the equivalence (24).

### 6.2 Hierarchies and Networks of Control

Of course, there is no reason why one should stop at *second*-order control. One could extend the setup with a finite *hierarchy* of control levels, with each level $u > 1$ defining who can transfer control of variables at level $u - 1$, and level $u = 1$ defining who can exercise first-order control. We then need to extend atomic programs to indicate which level of control is being transferred. Atomic programs then take the form:

$$i \leadsto_p^u j$$

to mean that agent $i$ transfers level $u$ control to agent $j$. The semantics of the language become yet more involved, but are straightforward to define. Somewhat related ideas were studied by Boella and van der Torre (2008).

Another direction is to consider multiple agents having 'write access' to propositional variables. For example, we might consider an 'authority relation' $\mathcal{P} \subseteq \mathbb{A} \times \mathbb{A}$, with the intended interpretation that $(i, j) \in \mathcal{P}$ means that everything that $i$ is empowered to everything that $j$ is empowered to do. Propositional variables are then allocated to sink nodes in $\mathcal{P}$ (i.e., agents with no outgoing edges in $\mathcal{P}$). One might then ask, for example, whether structural properties of the graph $\mathcal{P}$ characterise formulae of the object language.

## 7. Related Work

Although other researchers have begun to develop formal systems for reasoning about delegation and the transfer of control (e.g., Li, Grosof, & Feigenbaum, 2003), to the best of our knowledge DCL-PC is the first such system to have a rigorous semantics, and a complete axiomatization. Also, the emphasis of Li et al. (2003) is on decentralized 'trust management', in which roles like that of a requester, credentials and an authorizer are distinguished. In the work presented here, the emphasis is more on what coalitions can achieve, if they are allowed to hand over control over propositional variables.

Norman and Reed (2002) consider a logic of delegation, particularly focussing on group delegation. The logic underpinning this work is a 'STIT' (sees to it that) logic, in which the main operator is of the form $\mathsf{S}_i A$, meaning 'agent $i$ sees to it that $A$'. This extends to delegation by considering expressions of the form $\mathsf{S}_i \mathsf{S}_j$ ($i$ sees to it that $j$ sees to it that ...'). For example, an axiom in the resulting system is:

$$\mathsf{S}_i \mathsf{S}_j A \rightarrow \mathsf{S}_i A.$$

The work of Norman and Reed represents a serious attempt to develop a philosophically robust logic for delegation, appropriate for use in computational systems. However, their notion of delegation is very different to ours, (crudely, agents delegate responsibility, rather than transfer control), and the dynamic logic flavour of DCL-PC is absent. Finally, relatively few technical results relating to the logic are presented.

Jones and Sergot (1996) consider the problem of reasoning about power that an individual obtains by virtue of an organisational role. There, the notion of actions that are carried out in order to empower agents with certain capabilities is central, and Jones and Sergot also consider the interplay of such actions and ability. However, the logical formalisation is rather different – again a STIT-like language is used, rather than our dynamic logic framework, and relatively few technical results relating to the framework are presented. However, the setting of Jones and Sergot (1996) is much more general than ours: we focus only on propositional control. In somewhat related work, Boella and van der Torre (2006) present a formalisation of power delegation in the setting of *normative multi-agent systems*. They consider, for example, the issue of how delegated goals interact with other goals. The framework provides a rich and compelling setting for investigating questions relating to delegation. However, no overarching object language is developed for representing this framework, and relatively few technical results are presented relating to the framework. It would be interesting to consider whether the dynamic logic approach developed within the present paper might be adapted to the framework of Boella and van der Torre.

With respect to logics for reasoning about controlled variables, Boutilier (1994) presents a logic intended to capture notions such as 'I can achieve my plan using actions that only relate to variables under my control'. In spirit, this logic is very close to the kind of situation we are aiming to model, although the technical details of Boutilier's logic (the way control is captured in the logic) are different. Moreover, Boutilier's logic does not consider multi-agent aspects, or the dynamics of control as we do in the present paper.

We refer the reader to the work of van der Hoek and Wooldridge (2005b) for an extensive discussion and many references for logics of ability. Gerbrandy (2006) generalises the results of van der Hoek and Wooldridge, by considering situations in which an agent has only partial control of a variable, or where it shares control with others. Gerbrandy also shows how logics of propositional control are related to *cylindrical modal logic* (Venema, 1995). Specifically, the generalisation of CL-PC considered by Gerbrandy can be understood as a cylindrical modal logic, immediately yielding a complete axiomatization and decidability/undecidability results for various fragments of the system. A somewhat related formalism is discussed by van Benthem, Girard, and Roy (2009). This formalism is intended to enable reasoning about *ceteris paribus* preferences (in the sense of 'all other things being equal'). Van Benthem et al. develop a logic with a modality $\langle\Gamma\rangle\varphi$, where $\Gamma$ is a set of propositional formulae; the intended interpretation of $\langle\Gamma\rangle\varphi$ in a state $u$ is that there is a state $v$ agreeing with $u$ on the valuation of formulae $\Gamma$ in which $\varphi$ is true. There seems quite a close connection between DCL-PC and the formalism of van Benthem et al., although we leave the details for future work.

Our framework explains control in terms of *what* the agents can change or whcih *atoms* they can choose to be true or false. Sauro (2006) addresses the question of *how* agents can change the world, where control of coalitions is defined in terms of *actions* in the agents' repertoire. Finally, note that as discussed by van der Hoek and Wooldridge (2005b), the logic CL-PC is closely related to the well-known formalism of quantified Boolean formulae, and it is not hard to see that there is also a close relationship between DCL-PC and quantified Boolean formulae. However, while we may not ultimately have any gain is formal expressive power when using DCL-PC rather than quantified Boolean formulae, we *do* benefit with respect to the naturalness of expression in DCL-PC. Quantified Boolean formulae have no explicit notion of agency or the dynamics of control, and representing these aspects within quantified Boolean formulae leads to formulae that are unintuitive and hard to understand.

## 8. Conclusions

In this paper, we have built upon the logic CL-PC of strategic cooperative ability, in which the control that agents have over their environment is represented by assigning them specific propositional variables, for which the agents that 'own them' can determine their truth value. We added a dynamic component to this logic, thus obtaining the language DCL-PC in which one can reason about what agents (and coalitions of agents) can achieve by setting their assigned variables, or by giving the control over them to others. We gave two different but equivalent semantics for this language – a direct and a more conventional Kripke semantics – and provided a complete axiomatization for them. The key property that establishes the proof of completeness for DCL-PC's axiomatic system is the fact that every formula in the language is provably equivalent to a normal form: a disjunction of conjunctions of literals over propositional variables $p$ and assertions of the form *controls*$(i, p)$. We also investigated the complexity of the model checking and satisfiability problems for DCL-PC, and showed that these problems are no worse than for the program-free fragment CL-PC: they are both PSPACE-complete. We demonstrated that, for the special case where ability in ATL is interpreted as in (D)CL-PC, this implies a simpler satisfiability problem for ATL.

There are several avenues for further development of this work. First of all, it is interesting to add the assignments that the agents can perform to the transfer actions they can perform, so that the two dimensions of what agents can achieve become projected in one dimension. Although parallel execution is not a program construct in our language, and hence one could still not model situations where an agent chooses some values for its atoms, and *at the same time* transfers control of some other atoms, one could at least reason about the effect of programs that do a combination of truth assignments and transfer of control *in sequence*, or *in a choice*. Secondly, in many realistic systems, Property (22) may be too general: often, we want to specify that the overall system satisfies some constraints. For this, it seems appropriate, however, not only to reason about what agents *can* achieve, but also about what they *should* guarantee. The framework of *Social Laws* (Moses & Tennenholtz, 1995; van der Hoek, Roberts, & Wooldridge, 2005) could be set to work in order to express that under certain conditions, an agent will not set a certain propositional variable to 'true', or that she will not pass on control over a certain variable to a specific agent, or that the overall system behaves in such a way that every agent gets a fair chance to trigger a specific variable (i.e., use a specific resource) infinitely often. Another interesting direction would be to consider how to allow for the fact that agents outside a transfer program might change their variables while the program is executing. This might require some consideration of the semantics of parallel action. Relatedly, it would be interesting to make it possible to capture the temporal properties of the system, outside of transfer programs. Here, some combination of temporal and dynamic logic might be appropriate.

Similarly, we could weaken the *allocation* axiom to allow for some propositional not being under control of any agents, capturing the idea that not all facts are modifiable (by the agents under consideration). Another extension would be to assign control over atoms to *coalitions*, rather than *individual agents*. This could cater for 'power' in social contexts, with the typical example being that any coalition bigger than a threshold $n$ can lift a piano. Finally, an implementation of a theorem prover for the logic would of course be interesting.

**Acknowledgments** The authors wish to thank the JAIR reviewers and the editors for their useful comments. Michael Wooldridge and Dirk Walther were supported by the EPSRC under grant GR/S62727/01.

## Appendix A. Proofs

**Theorem 1.**

1. The schemes in Figure 6 are derivable in DCL-PC.
2. The axioms $K(i)$, $T(i)$, $B(i)$, and $\mathit{effect}(i)$ have coalitional counterparts $K(C)$, $T(C)$, $B(C)$, and $\mathit{effect}(C)$ that are all derivable for any coalition $C$.
3. $\vdash \mathit{controls}(C,p) \leftrightarrow \bigvee_{i\in C} \mathit{controls}(i,p)$.
4. The property $\mathit{persistence}_1(\mathit{control})$ is also derivable when we replace agent $i$ by an arbitrary coalition $C$.

**Proof:**

1. We describe how the eight schemes in Figure 6 can be derived in the axiomatic system of DCL-PC.
   - For $\mathit{at\text{-}least}(\mathit{control})$, this follows directly from axiom $\mathit{effect}(i)$, with taking $\psi = \top$.
   - For $\mathit{at\text{-}most}(\mathit{control})$, from $\ell(p)$ we get, using axiom $T(i)$ and contraposition, $\Diamond_i \ell(p)$. Assuming moreover $\Diamond_i \neg\ell(p)$, with axiom *control*, gives $\mathit{controls}(i,p)$. From *allocation* we then obtain $\neg\mathit{controls}(j,p)$ for any agent $j \neq i$, and, using $\mathit{control}(j)$, we get $\neg\Diamond_j p \vee \neg\Diamond_j \neg p$, i.e., $\neg\Diamond_j \ell(p) \vee \neg\Diamond_j \neg\ell(p)$. Since $T(j)$ gives us $\Diamond_j \ell(p)$, we obtain $\neg\Diamond_j \neg\ell(p)$, i.e., $\Box_j \ell(p)$.
   - To prove $\mathit{non\text{-}effect}(i)$, assume $\Diamond_i \ell(p) \wedge \neg\mathit{controls}(i,p)$. Then axiom $\mathit{control}(i)$ yields $\neg\Diamond_i \neg\ell(p)$, which is equivalent to $\Box_i \ell(p)$.
   - For $\mathit{persistence}(\mathit{non\text{-}control})$, the right-to-left direction follows immediately from $T(j)$. For the left-to-right direction, assume that $\neg\mathit{controls}(i,p)$. From *allocation* we derive that

     $$\mathit{controls}(1,p) \triangledown \cdots \triangledown \mathit{controls}(i-1,p) \triangledown \mathit{controls}(i+1,p) \triangledown \cdots \triangledown \mathit{controls}(n,p),$$

     and from this, by $\mathit{persistence}_1(\mathit{control})$ we get $\bigvee_{k\neq i} \Box_j \mathit{controls}(k,p)$. For every $k \neq i$, we have $\mathit{controls}(k,p) \rightarrow \neg\mathit{controls}(i,p)$, which follows from *allocation*. Hence, using *Necessitation*, we have $\Box_j(\mathit{controls}(k,p) \rightarrow \neg\mathit{controls}(i,p))$. From Axiom $K(j)$, it now follows $\Box_j \mathit{controls}(k,p) \rightarrow \Box_j \neg\mathit{controls}(i,p)$. Combining this with $\bigvee_{k\neq i} \Box_j \mathit{controls}(k,p)$, we obtain the desired conclusion $\Box_j \neg\mathit{controls}(i,p)$.
   - Notice that $\mathit{atomic\ permanence}(\leadsto)$ does not place any requirement on $p$, $i$ or $j$. Also, for the program $i \leadsto_p j$, under the condition that $\langle i \leadsto_p j\rangle\top$, by *func* and $\mathit{preconditions}(\mathit{transfer})$, $\langle i \leadsto_p j\rangle\varphi$ and $[i \leadsto_p j]\varphi$ are equivalent. Formally:

     $$\langle i \leadsto_p j\rangle\top \rightarrow (\langle i \leadsto_p j\rangle\varphi \leftrightarrow [i \leadsto_p j]\varphi) \tag{25}$$

Now, we prove *objective permanence*$(\rightsquigarrow)$ by induction on $\varphi$. The induction base, where $\varphi$ is a propositional variable, follows from *atomic permanence*$(\rightsquigarrow)$. Consider the induction step. Suppose the theorem is proven for $\psi$ and take $\varphi = \neg\psi$. Assume $\langle i \rightsquigarrow_p j\rangle\top$. It is to show that $\neg\psi \leftrightarrow [i \rightsquigarrow_p j]\neg\psi$ which is equivalent to showing that $\psi \leftrightarrow \langle i \rightsquigarrow_p j\rangle\psi$. This follows from (25) and the induction hypothesis. As a final step in the induction, suppose *objective permanence*$(\rightsquigarrow)$ is proven for $\psi_1$ and $\psi_2$. This means we can assume that

$$\langle i \rightsquigarrow_p j\rangle\top \rightarrow \big((\psi_1 \leftrightarrow [i \rightsquigarrow_p j]\psi_1) \wedge (\psi_2 \leftrightarrow [i \rightsquigarrow_p j]\psi_2)\big) \tag{26}$$

Now take $\varphi = \psi_1 \vee \psi_2$. Obviously, if we have $\psi_1 \vee \psi_2$, we have $[i \rightsquigarrow_p j](\psi_1 \vee \psi_2)$, which proves $\langle i \rightsquigarrow_p j\rangle\top \rightarrow (\varphi \rightarrow [i \rightsquigarrow_p j]\varphi)$. For the other direction, suppose, given $\langle i \rightsquigarrow_p j\rangle\top$, that $[i \rightsquigarrow_p j](\psi_1 \vee \psi_2)$. We use (25) to conclude $\langle i \rightsquigarrow_p j\rangle(\psi_1 \vee \psi_2)$, and by classical modal reasoning we then obtain $\langle i \rightsquigarrow_p j\rangle\psi_1 \vee \langle i \rightsquigarrow_p j\rangle\psi_2$. By (25) and the induction hypothesis we get $(\psi_1 \vee \psi_2)$, which concludes our proof.

- Using $K(\tau)$ and *Necessitation*, we can derive the fact $\neg\langle\tau\rangle\top \rightarrow [\tau]\varphi$. Using this fact, it is possible to show with propositional reasoning that *objective permanence* is equivalent to

$$[\tau]\varphi \leftrightarrow (\langle\tau\rangle\top \rightarrow \varphi) \tag{27}$$

The proof is by induction on the structure of the transfer program $\tau$. The first case of the induction base, where $\tau$ is an atomic program, holds by *atomic permanence*$(\rightsquigarrow)$. For the second case, suppose $\tau$ is a test $\psi?$. Notice that, by the axiom *test*$(\tau)$, $\langle\psi?\rangle\varphi$ is equivalent to $\psi \wedge \varphi$, and thus $\langle\psi?\rangle\top$ is equivalent to $\psi$. But then, (27) is equivalent *test*$(\tau)$.

Consider the induction step where $\tau = \tau_1;\tau_2$. The induction hypothesis tells us $\langle\tau_i\rangle\top \rightarrow ([\tau_i]\varphi \leftrightarrow \varphi)$, for all objective $\varphi$ and $i = 1, 2$. Assume $\langle\tau_1;\tau_2\rangle\top$; this implies $\langle\tau_1\rangle\langle\tau_2\rangle\top$ and $\langle\tau_1\rangle\top$ by *comp*$(\tau)$, and, by the induction hypothesis for $\tau_1$, $([\tau_1]\varphi \leftrightarrow \varphi)$. For any diamond operator, and hence also for $\langle\tau_1\rangle$, we have that if an implication $\psi \rightarrow \psi'$ is derivable, we can also derive $\langle\tau_1\rangle\psi \rightarrow \langle\tau_1\rangle\psi'$ using *Necessitation* and $K(\tau)$. Applying this to the induction hypothesis for $\tau_2$, i.e., to $\langle\tau_2\rangle\top \rightarrow ([\tau_2]\varphi \leftrightarrow \varphi)$, we obtain $\langle\tau_1\rangle\langle\tau_2\rangle\top \rightarrow \langle\tau_1\rangle([\tau_2]\varphi \leftrightarrow \varphi)$, and, with Modus Ponens, we arrive at $\langle\tau_1\rangle([\tau_2]\varphi \leftrightarrow \varphi)$. We like to demonstrate that $[\tau_1;\tau_2]\varphi \leftrightarrow \varphi$. By *comp*$(\tau)$, this is equivalent to $[\tau_1][\tau_2]\varphi \leftrightarrow \varphi$. For the direction from left to right, assume $[\tau_1][\tau_2]\varphi$. It is again a modal principle to conclude $\langle\tau_1\rangle\psi'$ from $\langle\tau_1\rangle(\psi \leftrightarrow \psi')$ and $[\tau_1]\psi$. Taking $\psi = [\tau_2]\varphi$ and $\psi' = \varphi$, we obtain $\langle\tau_1\rangle\varphi$. To show that $\varphi$ holds, suppose that $\neg\varphi$. This is still an objective formula, so we can apply the induction hypothesis to conclude $[\tau_1]\neg\varphi$. This, of course, contradicts $\langle\tau_1\rangle\varphi$, so that indeed we conclude $\varphi$. Conversely, suppose $\varphi$. Then, by the induction hypothesis for $\tau_1$, we also have $[\tau_1]\varphi$. The induction hypothesis for $\tau_2$, $\langle\tau_2\rangle \rightarrow ([\tau_2]\varphi \leftrightarrow \varphi)$ implies $\varphi \rightarrow [\tau_2]\varphi$, and we can apply necessitation and $K(\tau_1)$ to this to derive $[\tau_1]\varphi \rightarrow [\tau_1][\tau_2]\varphi$.

Now, consider $\tau = \tau_1 \cup \tau_2$, and *objective permanence* proven for $\tau_1$ and $\tau_2$. By axiom *union*$(\tau)$, we have $\langle\tau_1 \cup \tau_2\rangle\top \leftrightarrow (\langle\tau_1\rangle\top \wedge \langle\tau_2\rangle\top)$. But then, given $\langle\tau_1 \cup \tau_2\rangle\top$, we have: $[\tau_1 \cup \tau_2]\varphi \leftrightarrow ([\tau_1]\varphi \wedge [\tau_2]\varphi)$, and the induction hypothesis explains why the right-hand side of this equivalence is equivalent to $\varphi$.

Finally, consider $\tau = \tau_1^*$. By the axiom *mix*$(\tau)$, we immediately have $\langle\tau_1^*\rangle\top \rightarrow ([\tau_1^*]\varphi \rightarrow \varphi)$. For the other direction, recall that from the induction hypothesis we can derive, as a validity, $\varphi \rightarrow [\tau_1]\varphi$. Using *Necessitation* for $[\tau_1^*]$ then gives $[\tau_1^*](\varphi \rightarrow [\tau_1]\varphi)$. But then, using the assumption $\varphi$ and the axiom *ind*$(\tau)$ gives us $[\tau_1^*]\varphi$, hence we also have $\langle\tau_1^*\rangle\top \rightarrow (\varphi \rightarrow [\tau_1^*]\varphi)$.

- For axiom *inverse*, we rely on the normal form obtained in Theorem 2 (the proof of which does not involve *inverse*). So we know that every $\varphi$ is equivalent to a disjunction of formulas of the form $\varphi_1 \wedge \varphi_2$, where $\varphi_1$ is an objective formula, and $\varphi_2$ a conjunction of formulas of the form $controls(h,q)$. We now show that both $\varphi_1$ and $\varphi_2$ satisfy *inverse*, from which the result follows for arbitrary $\varphi$. So assume $controls(i,p)$. By $precondition(transfer)$ this entails $\langle i \rightsquigarrow_p j\rangle\top$, and hence we can apply $objectivepermanence(\rightsquigarrow)$ and *func* twice to conclude $\varphi_1 \leftrightarrow [i \rightsquigarrow_p j; j \rightsquigarrow_p i]\varphi_1$.

  Now we consider formulas $\varphi_2$, starting by their base case $controls(h,q)$. Assume $controls(i,p)$: we now first show the left to right direction. If $p \neq q$, we get from $persistence_2(control)$ that $[i \rightsquigarrow_p j; j \rightsquigarrow_p i]q$. If $p = q$ we consider three subcases: (1) $h = i$. We then derive, from $controls(i,p)$, using *transfer*, that $\langle i \rightsquigarrow_p j\rangle controls(j,p)$ and $controls(j,p) \rightarrow \langle j \rightsquigarrow_p i\rangle controls(i,p)$: with *func* and $comp(\tau)$, this gives $controls(i,p) \rightarrow [i \rightsquigarrow_p j; j \rightsquigarrow_p i]controls(i,p)$. (2) $h \neq i, h = j$. Given $controls(i,p)$, we then have $[i \rightsquigarrow_p j] \rightarrow \psi$ for any $\psi$, and we are done. (3) $h \neq i, h \neq j$. We can use $persistence_2(control)$ twice to get $controls(h,q) \rightarrow [i \rightsquigarrow_p j; j \rightsquigarrow_p i]controls(h,q)$. Finally, given $controls(i,p)$, we derive the right to left direction, that is, we derive $[i \rightsquigarrow_p j; j \rightsquigarrow_p i]controls(h,q) \rightarrow controls(h,q)$. First assume $p \neq q$ and suppose we would have $\neg controls(h,q)$. Then, by allocation, we have that for some agent $k \neq h$, we have $controls(k,q)$, and by $persistence_2(control)$ we get $[i \rightsquigarrow_p j; j \rightsquigarrow_p i]controls(k,q)$, which clearly contradicts $[i \rightsquigarrow_p j; j \rightsquigarrow_p i]controls(h,q)$. Now suppose $p = q$. Again, we have three subcases. (1) If $h = i$, the conclusion follows from the overall assumption $controls(i,p)$. (2) Suppose $h \neq i, h \neq j$. The same reasoning applies as in case (1). Finally, (3) suppose $h \neq i, h = j$. Since $controls(i,p)$ is given, we have $\langle i \rightsquigarrow_p j\rangle controls(j,p)$ (by *transfer*), and hence $\langle i \rightsquigarrow_p j; j \rightsquigarrow_p i\rangle controls(i,p)$. Now, if we would have $[i \rightsquigarrow_p j; j \rightsquigarrow_p i]controls(j,p)$, with $i \neq j$, this leads to a contradiction (use *allocation* and the fact that $h \neq i, h = j$), so indeed we derive $[i \rightsquigarrow_p j; j \rightsquigarrow_p i]controls(j,p) \rightarrow controls(j,p)$.

- For *reverse*, similar to inverse.

2. This was proved by van der Hoek and Wooldridge (2005b).

3. The definition of $controls(C,p)$ is $\Diamond_C p \wedge \Diamond_C \neg p$. Let $C = \{a_1, a_2, \ldots, a_C\}$. By axiom $control(i)$, we have $\bigvee_{i\in C} controls(i,p) \rightarrow \bigvee_{i\in C}(\Diamond_i p \wedge \Diamond_i \neg p)$. By the contrapositive of $T(i)$, we have $\varphi \rightarrow \Diamond_i\varphi$. We can apply this repeatedly for all agents in $C$, giving $\varphi \rightarrow \Diamond_{a_1}\Diamond_{a_2}\cdots\Diamond_{a_C}\varphi$. This is, according to *Comp*-$\cup$, the same as $\varphi \rightarrow \Diamond_C\varphi$. (Note that we have now proven the contrapositive of $T(C)$.) This gives us $\bigvee_{i\in C} controls(i,p) \rightarrow \bigvee_{i\in C}(\Diamond_C\Diamond_i p \wedge \Diamond_C\Diamond_i \neg p)$. Using *Comp*-$\cup$ again, we see that the consequent of this implication is equivalent to $\Diamond_C p \wedge \Diamond_C \neg p$. For the other direction, we first show $at-most(control)$ of Figure 6. From $\ell(p)$ we get, using axiom $T(i)$ and contraposition, $\Diamond_i\ell(p)$. Assuming moreover $\Diamond_i\neg\ell(p)$, with axiom $control(i)$, gives $controls(i,p)$. From *allocation* we then obtain $\neg controls(j,p)$, for any agent $j \neq i$. Using $control(j)$, we get $\neg\Diamond_j p \vee \neg\Diamond_j\neg p$, i.e., $\neg\Diamond_j \ell(p) \vee \neg\Diamond_j\neg\ell(p)$. Since $T(j)$ gives us $\Diamond_j \ell(p)$, we obtain $\neg\Diamond_j\neg\ell(p)$, i.e., $\Box_j \ell(p)$.

Now suppose $\neg \bigvee_{i\in C} controls(i,p)$. By *allocation*, we have $\bigvee_{x\in\mathbb{A}\setminus C} controls(x,p)$. That means that for one such $x$, we have $\Diamond_x p \wedge \Diamond_x\neg p$. Now we do a case distinction based on $p \vee \neg p$. In the first case, we assume $p$, and derive, for all $i \in C$, that $\Box_i p$, and thus $\Box_C p$. Hence $\neg\Diamond_C\neg p$, from which we get $\neg controls(C,p)$. In the case of $\neg p$, we similarly have, for all $i \in C$, that $\Box_i\neg p$, which gives $\neg\Diamond_C\neg\neg p$, and again $\neg controls(C,p)$. All in all, no matter whether $p$ or $\neg p$, we get $\neg controls(C,p)$.

4. This is easy: the previous item showed that $controls(C,p)$ means that $\bigvee_{i\in C} controls(i,p)$. Applying $persistence_1(control)$, we get $\bigvee_{i\in C} \Box_j controls(i,p)$. But since $controls(i,p) \rightarrow controls(C,p)$ if $i \in C$, we also have, for any $i \in C$, that $\Box_j controls(i,p) \rightarrow \Box_j controls(C,p)$ (use *Necessitation* and $K(j)$). This proves $\Box_j controls(C,p)$.

QED

## References

Alur, R., Henzinger, T. A., & Kupferman, O. (2002). Alternating-time temporal logic. *Journal of the ACM*, *49*(5), 672–713.

Blackburn, P., de Rijke, M., & Venema, Y. (2001). *Modal Logic*. Cambridge University Press: Cambridge, England.

Boella, G., & van der Torre, L. (2006). Delegation of power in normative multiagent systems. In *Deontic Logic and Artificial Normative Systems, 8th International Workshop on Deontic Logic in Computer Science, DEON 2006*, Utrecht, The Netherlands.

Boella, G., & van der Torre, L. (2008). Institutions with a hierarchy of authorities in distributed dynamic environments. *Artificial Intelligence and Law*, *16*(1), 53–71.

Boutilier, C. (1994). Toward a logic for qualitative decision theory. In *Proceedings of Knowledge Representation and Reasoning (KR&R-94)*, pp. 75–86.

Chellas, B. (1980). *Modal Logic: An Introduction*. Cambridge University Press: Cambridge, England.

French, T. (2006). *Bisimulation Quantifiers for Modal Logic*. Ph.D. thesis, The University of Western Australia, Perth, Australia.

Gerbrandy, J. (2006). Logics of propositional control. In *Proceedings of the Fifth International Joint Conference on Autonomous Agents and Multiagent Systems (AAMAS-2006)*, pp. 193–200, Hakodate, Japan.

Ghilardi, S., & Zawadowski, M. (2000). From bisimulation quantifiers to classifying toposes. In Wolter, F., Wansing, H., de Rijke, M., & Zakharyaschev, M. (Eds.), *Advances in Modal Logic*, pp. 193–220.

Goranko, V., & Jamroga, W. (2004). Comparing semantics of logics for multi-agent systems. *Synthese*, *139*(2), 241–280. In section *Knowledge, Rationality and Action*.

Harel, D., Kozen, D., & Tiuryn, J. (2000). *Dynamic Logic*. The MIT Press: Cambridge, MA.

Jamroga, W., & van der Hoek, W. (2004). Agents that know how to play. *Fundamenta Informaticae*, *63*(2-3), 185–219.

Jones, A. J. I., & Sergot, M. (1996). A formal characterisation of institutionalised power. *Logic Journal of the IGPL*, *3*, 427443.

Li, N., Grosof, B. N., & Feigenbaum, J. (2003). Delegation logic: A logic-based approach to distributed authorization. *ACM Transactions on Information and System Security*, *6*(1), 128 – 171.

Moses, Y., & Tennenholtz, M. (1995). Artificial social systems. *Computers and AI*, *14*(6), 533–562.

Norman, T. J., & Reed, C. (2002). Group delegation and responsibility. In *Proceedings of the First International Joint Conference on Autonomous Agents and Multiagent Systems (AAMAS-2002)*, pp. 491–498, Bologna, Italy.

Pauly, M. (2001). *Logic for Social Software*. Ph.D. thesis, University of Amsterdam. ILLC Dissertation Series 2001-10.

Sauro, L. (2006). *Formalizing Admissibility Criteria in Coalition Formation among Goal Directed Agents*. Ph.D. thesis, University of Turin, Turin, Italy.

Savitch, W. J. (1970). Relationships between nondeterministic and deterministic tape complexities. *Journal of Computer and Systems Sciences*, *4*(2), 177–192.

van Benthem, J., Girard, P., & Roy, O. (2009). Everything else being equal: A modal logic for ceteris paribus preferences. *Journal of Philosophical Logic*, *38*, 83125.

van der Hoek, W., Roberts, M., & Wooldridge, M. (2005). Knowledge and social laws. In Dignum, F., Dignum, V., Koenig, S., Kraus, S., Singh, M., & Wooldridge, M. (Eds.), *Proceedings of the Fourth International Joint Conference on Autonomous Agents and Multi-Agent Systems (AAMAS 05)*, pp. 674–681, New York, USA. ACM Inc.

van der Hoek, W., & Wooldridge, M. (2003). Time, knowledge, and cooperation: Alternating-time temporal epistemic logic and its applications. *Studia Logica*, *75*(1), 125–157.

van der Hoek, W., & Wooldridge, M. (2005a). On the dynamics of delegation, cooperation, and control: A logical account. In Dignum, F., Dignum, V., Koenig, S., Kraus, S., Singh, M., & Wooldridge, M. (Eds.), *Proceedings of the Fourth International Joint Conference on Autonomous Agents and Multi-Agent Systems (AAMAS 05)*, pp. 701–708, New York, USA. ACM Inc.

van der Hoek, W., & Wooldridge, M. (2005b). On the logic of cooperation and propositional control. *Artificial Intelligence*, *64*, 81–119.

Venema, Y. (1995). Cylindric modal logic. *Journal of Symbolic Logic*, *60*, 591623.